\documentclass[journal]{IEEEtran}
\ifCLASSINFOpdf
\else
\fi
\usepackage{graphicx}
\usepackage{subfigure}
\usepackage{booktabs}
 \usepackage{amsmath}

\begin{document}
%
\title{Impact Mitigation for Dynamic Legged Robots with Steel Wire Transmission Using Nonlinear Active Compliance Control}
%
%
%

\author{Junjie Yang,~\IEEEmembership{Member,~IEEE,}
        Hao Sun,~\IEEEmembership{Member,~IEEE,}
        Hao An,~\IEEEmembership{Member,~IEEE,}
        and~Changhong~Wang,~\IEEEmembership{Senior~Member,~IEEE}
\thanks{The authors are with Space Control and Inertial Technology Research Center, Harbin Institute of Technology, Harbin, 150001, P.R. China(e-mail:cwang@hit.edu.cn).}}

%
%

\markboth{IEEE ROBOTICS AND AUTOMATION LETTERS,}%
{Shell \MakeLowercase{\textit{et al.}}: Bare Demo of IEEEtran.cls for IEEE Journals}
%



\maketitle

\begin{abstract}

Impact mitigation is crucial to the stable locomotion of legged robots, especially in high-speed dynamic locomotion. This paper presents a leg locomotion system including the nonlinear active compliance control and the active impedance control for the steel wire transmission-based legged robot. The developed control system enables high-speed dynamic locomotion with excellent impact mitigation and leg position tracking performance, where three strategies are applied. $a)$ The feed-forward controller is designed according to the linear motor-leg model with the information of Coulomb friction and viscous friction. $b)$ Steel wire transmission model-based compensation guarantees ideal virtual spring compliance characteristics. $c)$ Nonlinear active compliance control and active impedance control ensure better impact mitigation performance than linear scheme and guarantees position tracking performance. The proposed control system is verified on a real robot named SCIT Dog and the experiment demonstrates the ideal impact mitigation ability in high-speed dynamic locomotion without any passive spring mechanism.

\end{abstract}


%
\IEEEpeerreviewmaketitle

 \small
\section{Introduction}
%
%
%
%
\IEEEPARstart{T}{he} animals such as dogs and horses have a strong locomotion ability in the field. Based on bionics, legged robots can also perform complex field tasks and flexible motions, theoretically. Traditionally, most robots including legged robots and manipulators are driven by rigid position control with large gearbox transmissions and high-gain controllers \cite{Hirai2002Current}, which results stiff position-controlled joints. The method of generating free gaits \cite{Estremera2005Generating} gives legged robots the ability of static locomotion in unstructured terrains, which, however, is not suitable for dynamic locomotion. When a leg of the robot interacts with the ground, it will be forced to execute the locomotion according to the original position control signal, causing a huge impact on the leg and potentially leading to unstable locomotion of the robot. Although legged robots controlled by position also realize the basic walking, they cannot achieve high-speed dynamic running trot or bounding. Conversely, the legs of real animals have natural cushioning structures, such as tendons and plantar pads. Moreover, animals can also bend knees through nerve feedback to realize impact mitigation, this control naturally has a delay of few tens of milliseconds or more \cite{2013Kandel}\cite{2010Geyer}.

Since the trajectory of the leg locomotion of a robot is generally predefined in the swing phase, rigid position control cannot be effectively applied in a non-ideal environment with unstructured terrains. A strong ability to continuously and safely interact with the real environment like animals usually requires modelling the surrounding environment and estimating robot states exactly, which, however, is difficult for legged robots to realize. Aiming at mitigating the impact on legs, animals have the passive impact mitigation mechanism as well as the ability to actively regulating muscle compliance and impedance \cite{Burdet2001The}. Likewise, Hogan et al. \cite{Hogan1984Adaptive} found that the adjustment of impedance is a key factor for the robot to walk on unstructured terrains. Meanwhile, \cite{Hogan2009Impedance} showed that controlling contact force and joint torque is crucial. For the interaction between legs and ground, Raibert \cite{Raibert1986Legged} proposed a spring passive dynamics method based on the spring-loaded inverted pendulum model; Pratt et al. \cite{Williamson1995Series} proposed the series elastic actuator (SEA). SEA connects springs in series with the actuators, so as to absorb impact and release energy in the next phase \cite{Laffranchi2014Development}. This method protects the structure of the robot from impacts while saving energy. However, passive springs cause legs to be unable to complete specific rigid locomotion, in which case the position tracking performance becomes worse. Moreover, the series spring has a natural vibration frequency that limits the bandwidth of force control and requires additional tuning of the system. In order to solve the invariable stiffness of passive spring, variable stiffness actuator (VSA) is a relatively novel solution, where an additional actuator adjusts spring stiffness. Tsagarakis et al. \cite{Tsagarakis2011A} proposed a new VSA design to increase the range of adjustable stiffness. Kim et al. \cite{Kim2012Design} designed a hybrid variable stiffness actuator (HVSA), achieving a wide range of stiffness and fast response. Unfortunately, VSAs are still bulky, complex and often cannot absorb high-energy impacts due to the limited size of the springs \cite{Lin2018Experimental}\cite{Sun2018}\cite{Lin2020Low}.

Interacting with the unstructured environment without mechanical springs is crucial for legged robots. By conducting a reasonable design of the control system, the mechanical compliance and impedance of the joint can be adjusted alone in the software, resulting in a virtual spring-damper on the robot leg (called active impedance/compliance control). To realize the functions of the virtual spring-damper through software, especially in high dynamic performance, the fast response of the control system is crucial. As a result, a sufficient torque control frequency is necessary to effectively replace a real spring with a virtual spring for legged robots. Virtual springs simplify the design of robot mechanisms. At present, active compliance control has been widely implemented in the field of legged robots, which enables many legged robots to interact with the ground without any physical springs. Wensing et al. implemented impedance control to build the well-known cheetah robots \cite{Wensing2017Proprioceptive} \cite{Seok2012Actuator}. Hyun et al. \cite{Dong2014High} used a low transmission ratio of proprioceptive impedance torque control mode, achieving high-speed locomotion on the treadmill, and the Froude number reached 7.1. Semini et al. \cite{Semini2016} implemented active impedance control on hydraulic actuators and realized efficient flying trot on the 80 kg HYQ robot. HYQ robot has a good impact mitigation performance and position tracking performance. Hutter et al. \cite{Hutter2013Efficient} designed high-efficiency and energy-saving robotic legs by combining passive compliance with active compliance and manufactures StarlETH robot by implementing operation space control. These results show that the active compliance/impedance control can be successfully implemented on legged robots. Note that the researches mentioned above only considered linear compliance, which can not fill the demand for flexibly balancing between stable locomotion and impact mitigation according to different robot states. Therefore, nonlinear compliance should be considered to achieve a satisfying trade-off between the two.

In this paper, we present a nonlinear active compliance/impedance control and design a leg locomotion control scheme based on the presented control method. Nonlinear active compliance provides a better impact mitigation performance than the existing linear scheme, which will be demonstrated in experiments. The main contribution of this paper is the design of a leg locomotion control scheme, with purely active compliance/impedance controlled legs and no mechanical springs. Nonlinear active compliance and impedance control are implemented on a real robot named SCIT Dog. The limitation of the stability of the closed-loop control system to the time-varying parameters of the controller is also derived. Compared with the linear scheme, it has a better impact mitigation performance. To the best of the authors’ knowledge, this is the first time that the nonlinear active compliance is implemented on quadruped robots. The steel wire transmission scheme is rarely applied to quadruped robots for reducing the inertia of the robot's legs. Other works aim to solve the friction interference and the steel wire transmission interference faced by the leg control system of SCIT Dog in practical application, including: 1) Coulomb friction and viscous friction are considered in the linear motor-leg model, and the feed-forward controller is designed based on the model. The position tracking performance and high-speed dynamic performance of the leg also benefit from the design scheme of the feed-forward controller. 2) A compensation method is proposed, which can effectively deal with the influence of high order dynamics in steel wire transmission system on the robot stable locomotion.

This paper is structured as follows: Section 2 introduces the structure design of SCIT Dog. The motor-leg model and friction model of SCIT Dog are established, systematically. Section 3 gives the overall leg torque control scheme of SCIT Dog based on central pattern generator (CPG) locomotion planning. The control framework based on the feed-forward control will be introduced firstly. Then, the PI torque controller for the motor-leg model and friction compensation will be introduced briefly. Finally, active compliance/impedance control and the influence of virtual spring and damping on foot-to-ground interaction will be discussed in detail, and the stability condition of the closed-loop system is derived. Section 4 is the experiment part. The position tracking experiments and impact experiments are carried out on the SCIT Dog. The impact mitigation performance and position tracking performance of the nonlinear active compliance/impedance control are demonstrated. Conclusions will be given in Section 5.


\section{Modeling of SCIT Dog}

The considered SCIT Dog is driven by 12 brushless DC motors and has 12 joint degrees of freedom (DoF) and 6 body DoFs. Each leg contains three joints, corresponding to the hip and knee joints of a real dog. Since the hip joint has two DoFs, namely, span and swing, the function of the hip joint should be realized by two motors. The DoF of the knee joint is driven by steel wire transmission, steel wire is wound around two fixed transmission wheels. One wheel is connected to the knee joint, and the other wheel is connected to the motor. The control motor for the knee is designed in the position of the hip joint, above the motor that controls the swing DoF of the hip joint (Figure 1 (a)). Only the degree of freedom of the knee joint is controlled by steel wire, and the two degrees of freedom of the hip joint are rigidly connected. (Although the load distribution of the robot leg joint is related to the joint configuration, the knee joint bears the main impact during the robot locomotion.) This design has two advantages as follows: (1) The motor designed in the hip joint drives the motion of the knee joint through steel wire, which greatly reduces the inertia of legs. (2) In some extreme cases, if the robot leg suffers from excessive force, the steel wire will break to protect other structures of the leg from being damaged. No passive springs are designed on SCIT Dog, and the desired compliance and impedance of the legs come from the control scheme. However, steel wire transmission causes additional problems to joint control, primarily the nonlinear relationship between the motor and joint velocity, which will be considered and compensated in Section 3. Each robot leg has three segments, which are thigh, leg, and foot respectively. The foot is connected with the thigh through connecting rods and keeps parallel with the thigh. This design can realize three segments of legs controlled by using only two motors. SCIT Dog is shown in Figure 1.

\begin{figure}[!ht]
    \centering
    \subfigure[SCIT Dog.]{
    \begin{minipage}[t]{0.22\textwidth}
    \centering
    \includegraphics[width=2.2in]{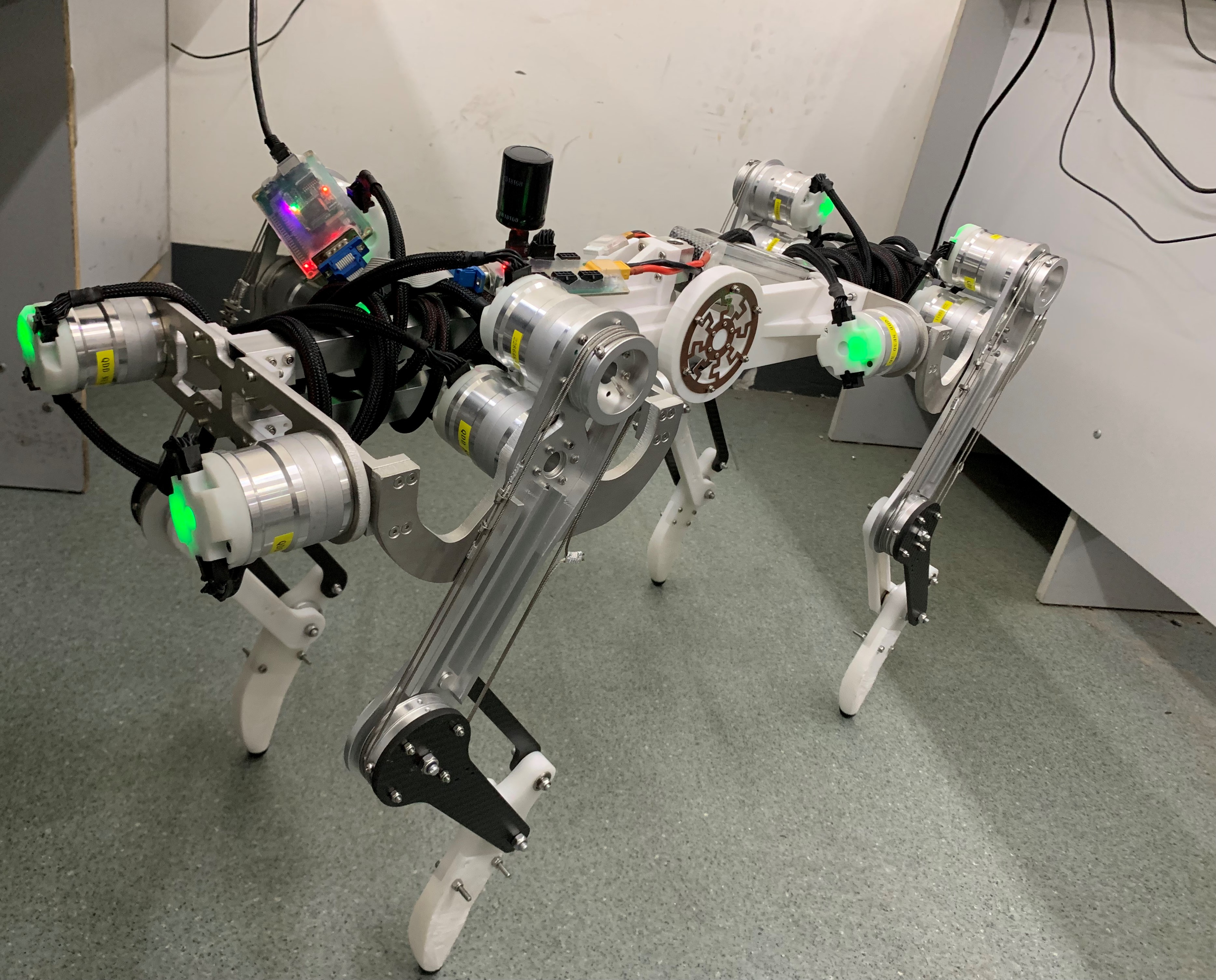}
    \end{minipage}
    }
    \subfigure[Design of leg.]{
    \begin{minipage}[t]{0.22\textwidth}
    \centering
    \includegraphics[width=1.3in]{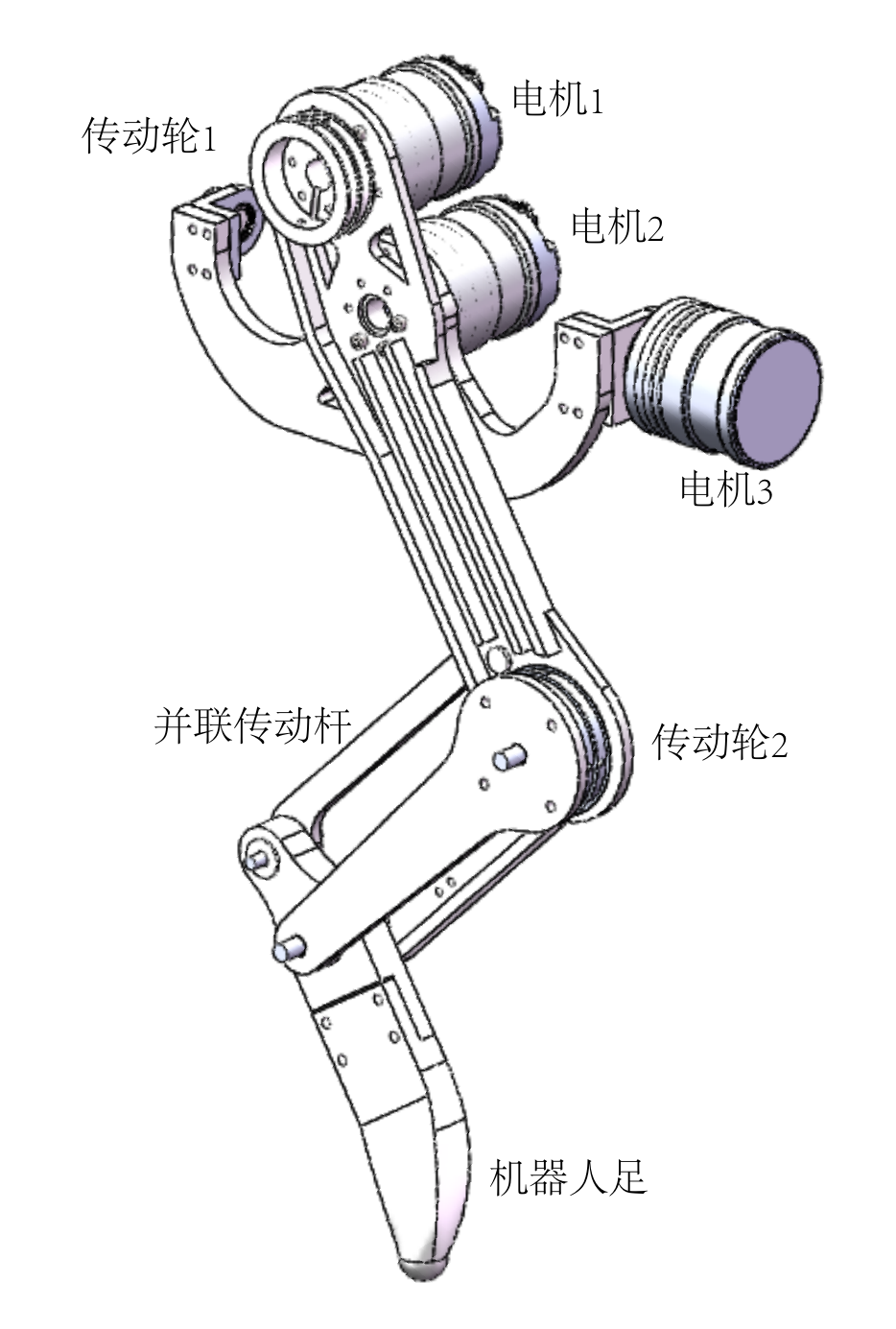}
    \end{minipage}
    }
    \caption{Illustration of the considered SCIT Dog. 12 motors are designed at the hip joint and the knee joints are transmitted by steel wire, which can effectively reduce the inertia of the leg.}
    \label{DOG}
\end{figure}

\subsection{Motor-leg model}

An accurate motor-leg model is critical to implementing active compliance and impedance control. Essentially, the motor-leg model is equivalent to a motor-load model. The input of the motor model is torque and the output is angular velocity. The input and output of the load model are all both angular velocities. The steel wire can be considered as a linear spring with large stiffness. Considering that SCIT dog's encoders are installed to measure the output of the reducer gearboxes, and the reducer ratio of the gearboxes can be ignored in the model. Therefore, the dynamic model of the load has natural velocity feedback \cite{Boaventura2017Model}, rather than a simple rigid model, which is formulated as

\begin{equation}
\dot{\tau_l}=K_w(\omega_m-\omega_l)
\end{equation}
where $\tau_l$ is the torque exerted on the load of the leg, $K_w$ is the stiffness coefficient of steel wire, $\omega_m$ is the angular velocity of the motor, and $\omega_l$ is the angular velocity of the crus. In fact, the torque that the external force maps to the joint space has an effect on $\tau_l$. The whole motor-leg model is shown in Figure 2.

\begin{figure}[!ht]
\centering
\includegraphics[width=3.5in]{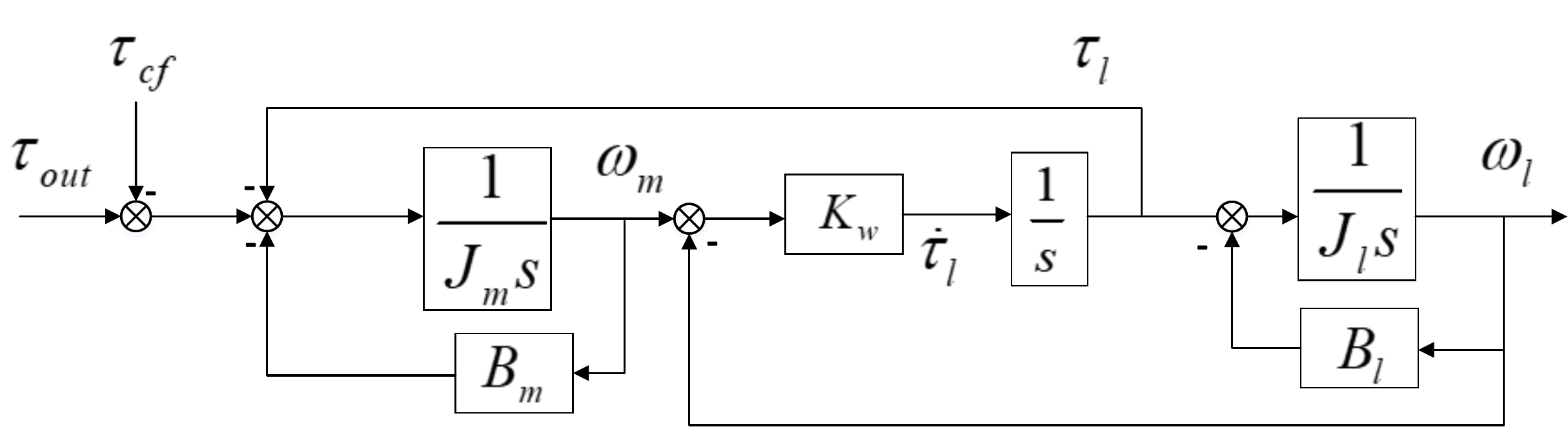}%
\caption{Motor-leg model diagram.}
\label{fig_sim}
\end{figure}

Further, define notations as follows. $\tau_{out}$ is the output torque of torque controller, $\tau_{cf}$ is the static Coulomb friction, $J_m$ is the inertia of the motor, $B_m$ is the viscosity coefficient of the motor, $J_l$ is the inertia of the load, $B_l$ is viscosity coefficient of the load. As a result, the transfer function of the model is obtained according to the diagram.

\begin{equation}
\frac{\omega_l(s)}{\tau_{out}(s)}=\frac{K_w}{(K_w+J_ls^2+B_ls)(J_ms+B_m)+K_w(J_ls+B_l)}
\end{equation}

The output torque of the motor is proportional to the current of the motor, while there is a complex electrical dynamic relationship between motor voltage and current. Therefore, a control method for motor current rather than voltage is implemented on SCIT dog to reduce the controller complexity. A PI controller is designed by implementing motor current feedback so that current can be directly applied, and torque can be directly applied to the input of the motor-leg model. This will be explained in detail in Section 3.

\subsection{Friction model and load torque}

Friction exists in both motor and load. Friction consists of two parts, Coulomb friction and viscous friction \cite{Tzes1998Genetic}. For motors, total friction can be expressed as

\begin{equation}
\tau_{fm}=\tau_{cfm}+B_m\omega_m
\end{equation}
where $\tau_{fm}$ is the total friction of the motor and $\tau_{cfm}$ is Coulomb friction of the motor. For loads, friction can also be expressed as

\begin{equation}
\tau_{fl}=\tau_{cfl}+B_l\omega_l
\end{equation}
where $\tau_{fl}$ is total friction of load and $\tau_{cfl}$ is Coulomb friction of load. So the total friction torque of the motor-leg model can be obtained.

\begin{equation}
\tau_{f}=\tau_{cfm}+B_m\omega_m+\tau_{cfl}+B_l\omega_l
\end{equation}

According to the motor-leg model, the actual torque applied to the load can be obtained.

\begin{equation}
\tau_{l}=\tau_{out}-\tau_{fm}-J_m\dot{\omega}_m
\end{equation}

The angular acceleration of the joint may be erroneous through simple numerical differentiation from an optical encoder signal \cite{2001Acceleration}. There is also such difficulty in measuring acceleration for SCIT Dog, so the joint angular acceleration is designed to be obtained by input based on the CPG locomotion signal, which avoids the measurement requirement. The generation of the reference trajectories for the feet is inspired by the CPGs of
animals \cite{2008Central}. The trajectories of CPG  has intuitive parameters such as step length and step height, instead of angular joint displacement.
The feed-forward torque control framework is implemented on SCIT dog, which efficiently applies the angular acceleration as input. The control framework will be explained in Section 3.

\section{Leg locomotion control framework}

For CPG signals, the trajectories of the robot feet are planned, and both the desired velocity and acceleration are contained in the planned trajectories. The desired position and velocity are implemented by active compliance and impedance torque control, while the desired acceleration of the robot system is implemented on SCIT Dog by feed-forward torque control. The overall control block diagram is shown in Figure 3.

\begin{figure}[!ht]
\centering
\includegraphics[width=3.5in]{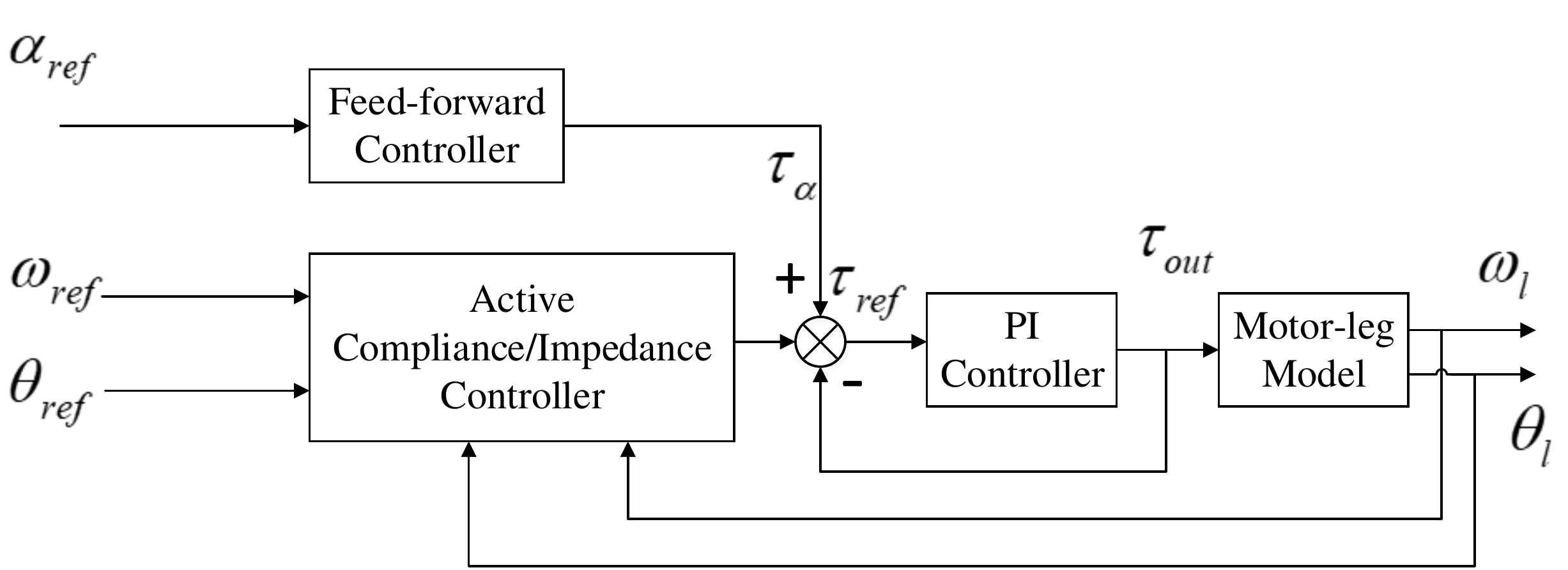}%
\caption{Leg locomotive control framework.}
\label{fig_sim}
\end{figure}

In Figure 3, $\alpha_{ref}$ is the desired angular acceleration, $\tau_{\alpha}$ is the output of the feed-forward controller, $\omega_{ref}$ and $\theta_{ref}$ are the reference joint angular and joint angle velocity calculated by the desired foot trajectory, $\tau_{ref}$ is the reference torque for the PI controller. Inputs of the control framework are $\alpha_{ref}$, $\omega_{ref}$ and $\theta_{ref}$, and outputs of the control framework are $\omega_{l}$ and $\theta_{l}$. The output of the feed-forward controller is the torque corresponding to the desired joint angular acceleration plus the total friction torque. Torque controllers are divided into the inner loop and outer loop. In the inner loop, the high-frequency PI torque controller can effectively control the output torque of the motor to follow the desired torque. Meanwhile, in the outer loop, the active compliance/impedance controller is the key to realize both the impact mitigation performance and position tracking performance. In the rest of this section, we will first introduce the design of the PI torque controller and the realization of the friction compensation method briefly, and then, take focus on the design of the proposed novel active compliance/impedance controller.

\subsection{Inner loop PI torque controller}

The direct input of the motor system is voltage, while the output torque is linear with the motor current. The motor voltage and current have a complex motor dynamic relationship, and voltage control commonly causes torque ripple. To guarantee the control frequency, a PI controller for current is implemented on SCIT Dog, which is simple and effective. Here, the feedback information of the controller is the current sensor. As we all know, the Gaussian noise of the current sensor is huge. For the commonly used PID controller, the differential term will reduce the performance of the controller. The integral term plays an important role in reducing the static error of the system, so compared with P, PD, PID controller, it is most reasonable to use PI controller for torque control. Although the characteristics of the PI controller produces non-ideal oscillation in the motor current. Nevertheless, the current control system is in the inner loop of the whole control system, and practical experiments demonstrate that the oscillation of the internal control loop has little effect on the external control loop. Behind the PI controller, a filter link and a saturation link are added. The internal loop of torque control for motors is expressed in Figure 4.

\begin{figure}[!ht]
\centering
\includegraphics[width=3.5in]{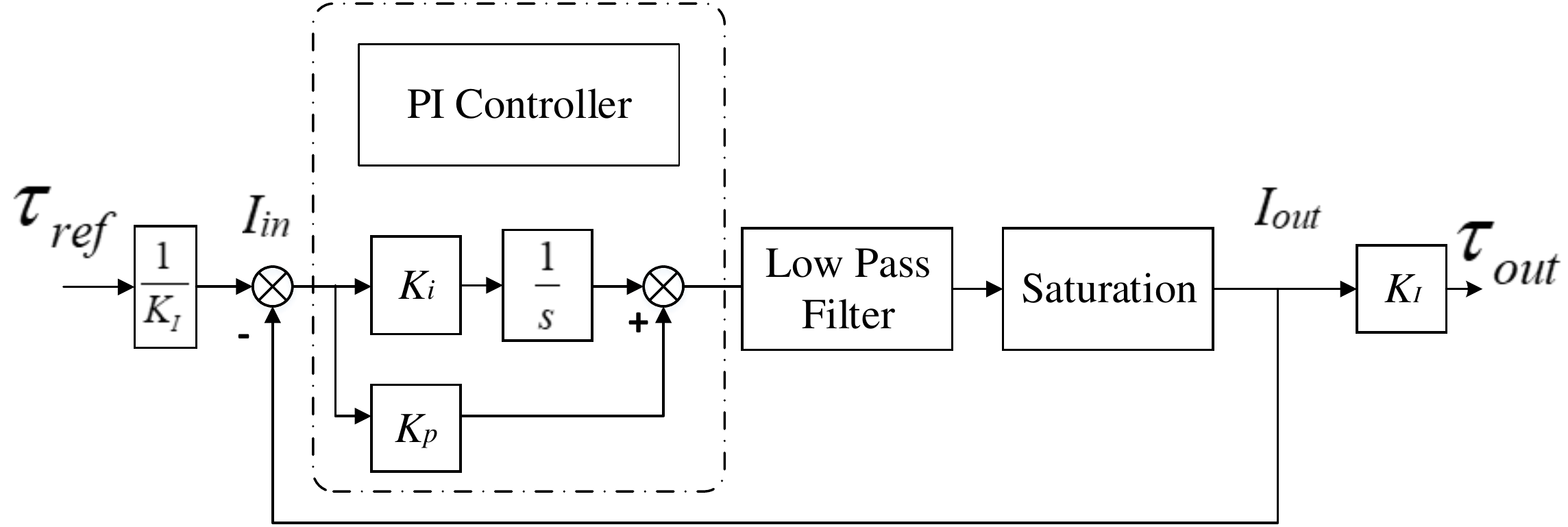}%
\caption{PI current controller.}
\label{fig_sim}
\end{figure}

Further, define notations as follows. $\tau_{out}$ is an input of the motor-leg system. $K_I$ is the torque coefficient of the motor. $I_{in}$ and $I_{out}$ are desired current and actual current, $K_p$ and $K_i$ are parameters of the PI controller.

\subsection{Friction torque compensation}

According to the friction torque model introduced above, the actual angular velocity of the motor and load is indispensable. The feed-forward controller is described by

\begin{equation}
\tau_\alpha=(J_m+J_l)\theta_{ref}+\tau_{cfm}+B_m\omega_m+\tau_{cfl}+B_l\omega_l+\tau_{cor}+\tau_g
\end{equation}
where $\tau_{cor}$ and $\tau_g$ are torque due to Coriolis and gravity respectively. The angular velocity of the motor is obtained directly from the encoder information. However, the actual angular velocity information of the load needs additional sensors. To avoid additional sensors, the dynamic model of the load can be used to estimate the load angular velocity. Because there are many nonlinear factors in the actual friction torque, friction compensation uses the estimated velocity of the load. The main reason for the nonlinear relationship between load velocity and motor velocity is caused by the compliance of the steel wire transmission, which is reflected in the knee joint. Therefore, for the hip joint, the motor angular velocity and load angular velocity is linear. According to the load model, the transfer function of load speed is obtained.

\begin{equation}
\frac{\omega_l(s)}{\omega_m(s)}=\frac{K_w}{K_w+J_ls^2+B_ls}
\end{equation}

The estimated load velocity is obtained by solving Laplace inverse transformation, and the input of the system is regarded as a fixed value in each control cycle. Because of the stiffness of the steel wire is relatively large, the nonlinear motor-load velocity difference will be generated only when the legs interact with the ground. Therefore, the nonlinear motor-load velocity difference generated in the stance phase will be compensated quickly in the swing phase. This characteristic makes the motor-load velocities to be linear at the beginning of each stance phase. The initial value of load velocity is calculated by motor velocity.

\subsection{Nonlinear active compliance and impedance torque control}

The impact between the feet and the ground is the most important issue discussed in this work, which may damage the structure of the leg and some fragile connections. Even if the structure is not damaged, it will become less compact after a long-term operation, enlarging the dead zone of the transmission. Such an impact can cause an undesirable shake in the center of mass (COM) of the robot, resulting in instability locomotion even leading the body to fall. To mitigate the impacts, robots must have the ability to control forces that interact with the ground, which requires adjustable compliance/impedance of the robot's leg joints. When quadruped animals run and walk, they can actively regulate leg compliance by changing the stiffness of leg muscles. Robots need to imitate this ability to actively adjust compliance so that the impact can be effectively mitigated, and compliance needs to be gradually reduced to ensure the accuracy of foot position tracking after the impact. There are two kinds of leg compliance, which are passive and active respectively. Passive compliance is achieved by adding flexible elements such as springs between actuators and loads. Active compliance means that compliance control is added to the robot's torque control algorithm, and the coefficients can be adjusted by software in real-time so that the robot leg has the same potential as the dog leg to control the interaction force. Active compliance torque is implemented by position and velocity feedback, which is described by

\begin{equation}
\boldsymbol{\tau}_{ref}=\boldsymbol{J}^T[\boldsymbol{K}_{vs,m}(\boldsymbol{x}_{des}-\boldsymbol{x})+\boldsymbol{K}_{d,m}(\dot{\boldsymbol{x}}_{des}-\dot{\boldsymbol{x}})]+\boldsymbol{\tau}_{\alpha}
\end{equation}
where $\boldsymbol{J}$ is the foot Jacobian matrix, $\boldsymbol{K}_{vs,m}$ $\in$ $R^{3\times3}$ is stiffness coefficient matrix, $\boldsymbol{K}_{d,m}$ $\in$ $R^{3\times3}$ is damping coefficient matrix, $\boldsymbol{x}_{des}$,$\boldsymbol{x}$ $\in$ $R^3$ is the expected and actual position of robot foot. Because compliance control does not need to guarantee zero steady-state error, the integral term $I$ is not needed in the controller.

This method is equivalent to realizing a virtual spring-damper on the robot leg. The proportional term can be regarded as a spring and the differential term can be regarded as a damper. $K_{vs}$ is the elastic coefficient of the virtual spring and $K_d$ is the damping coefficient of the damper. Note that $K_{vs}$ and $K_d$ are the parameters of the virtual spring, the diagonal terms of the stiffness and damping matrices $\boldsymbol{K}_{vs,m}$ and $\boldsymbol{K}_{d,m}$, respectively. The virtual spring-damper element is shown in Figure 5.

\begin{figure}[!ht]
\centering
\includegraphics[width=2.0in]{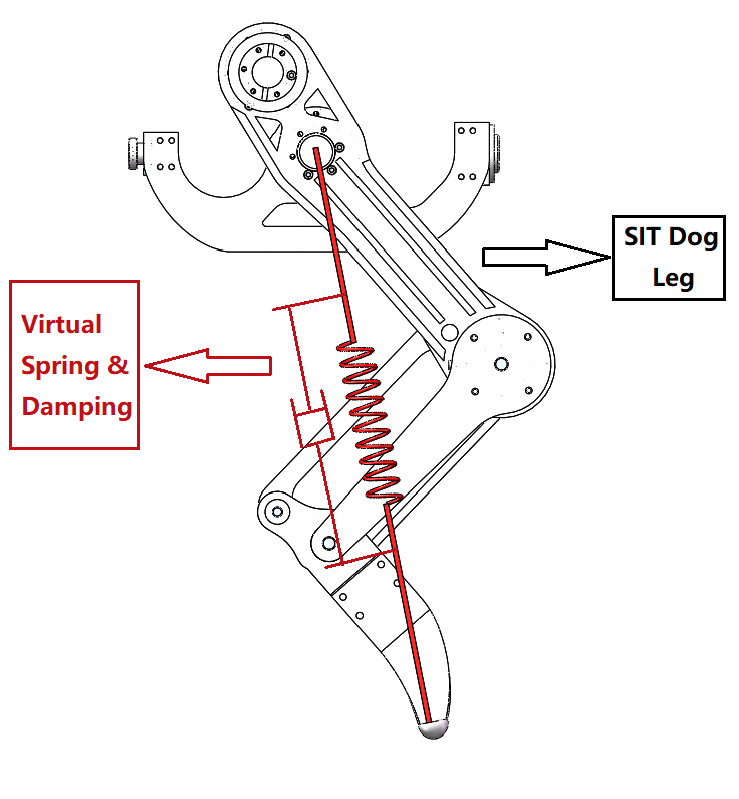}%
\caption{Leg and virtual spring-damper.}
\label{fig_sim}
\end{figure}

According to the motor-leg model, the transmission wire has non-negligible compliance, which will lead to additional interference in the design of $K_{vs}$ function. This disturbance means that there is another spring element in the designed virtual spring that is caused by the compliance of the steel wire so that the designed virtual spring is not an ideal functional characteristic. The transmission wire is compensated by fine-tuning the $K_{vs}$ curve. The nesting of two springs is not a simple linear superposition of two elastic coefficients. Assume that the springs with two stiffness coefficients $K_{des}$ and $K_w$ are in series, and the total stiffness coefficient is $K_{vs}$. This assumption is not perfect because the forces acting on the springs are not in the same direction. But in practice, this assumption is reasonable because the stiffness coefficient of steel wire cannot be accurately measured (the accuracy is the order of magnitude). The deformations of the springs with stiffness coefficients $K_{des}$ and $K_w$ are $\bigtriangleup x_1$ and $\bigtriangleup x_2$ respectively, and the total deformation is $\bigtriangleup x=\bigtriangleup x_1+\bigtriangleup x_2$. Through $K_{vs}\bigtriangleup x=K_{des}\bigtriangleup x_1=K_w\bigtriangleup x_2$, the compensated value of $K_{vs}$ is approximately described as

\begin{equation}
K_{vs}=\frac{K_{des}K_w}{K_w-K_{des}}
\end{equation}
where $K_{des}$ is the desired elastic coefficient. The denominator of $K_{vs}$ is always positive because the stiffness of the steel wire is much larger than the desired spring stiffness.

In the actual leg, the characteristics of the virtual spring have been verified as follows. SCIT Dog's single leg is manually compressed. Given the desired foot position, the position deviation can be obtained by joint angle and leg kinematics, and the output force is measured. The virtual spring characteristics of the legs are obtained by repeated experiments, which are shown in Figure 6.

\begin{figure}[!ht]
    \centering
    \subfigure[Linear spring characteristic.]{
    \begin{minipage}[b]{0.45\textwidth}
    \centering
    \includegraphics[width=2.5in]{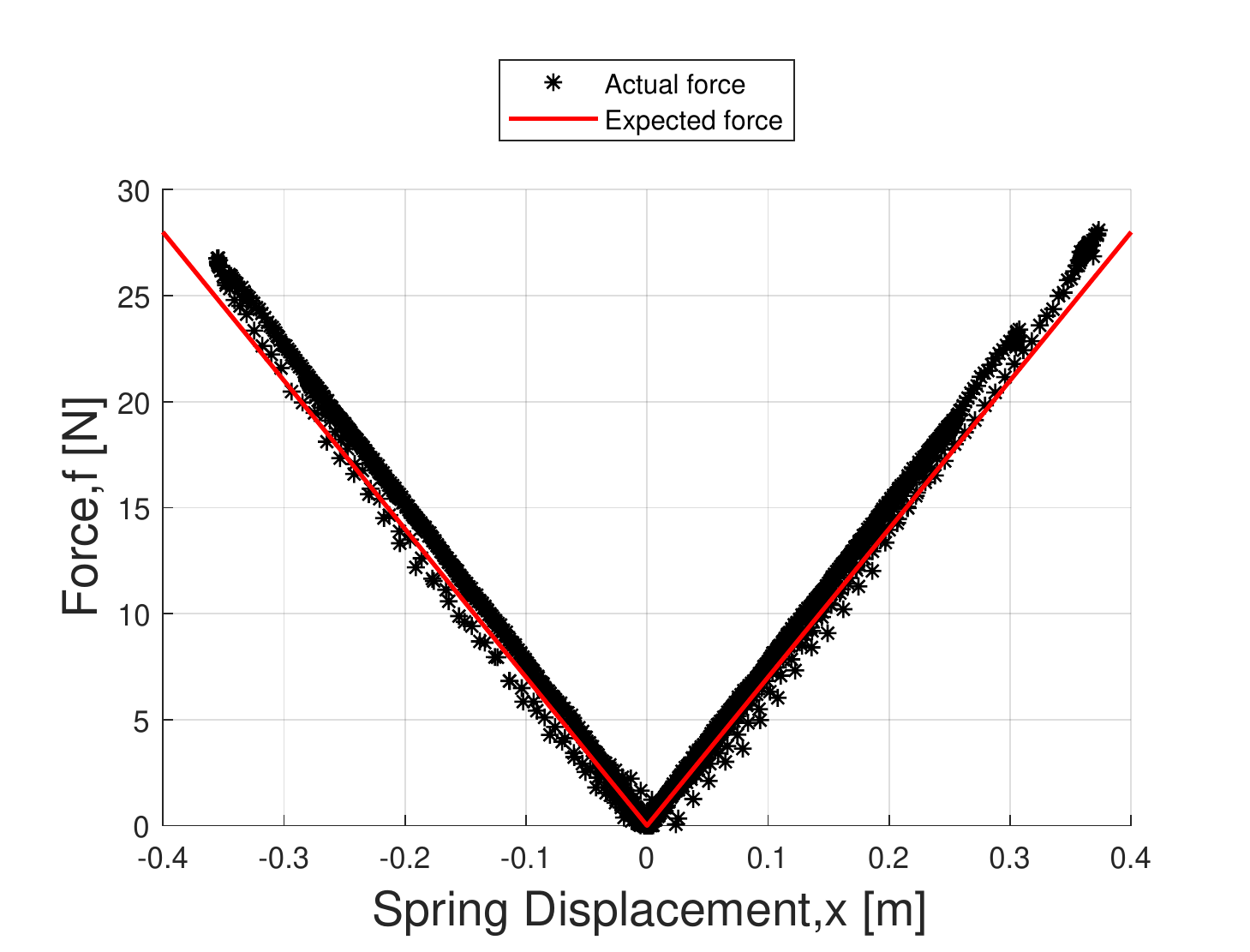}
    \end{minipage}
    }
    \subfigure[Nonlinear spring characteristic.]{
    \begin{minipage}[b]{0.45\textwidth}
    \centering
    \includegraphics[width=2.5in]{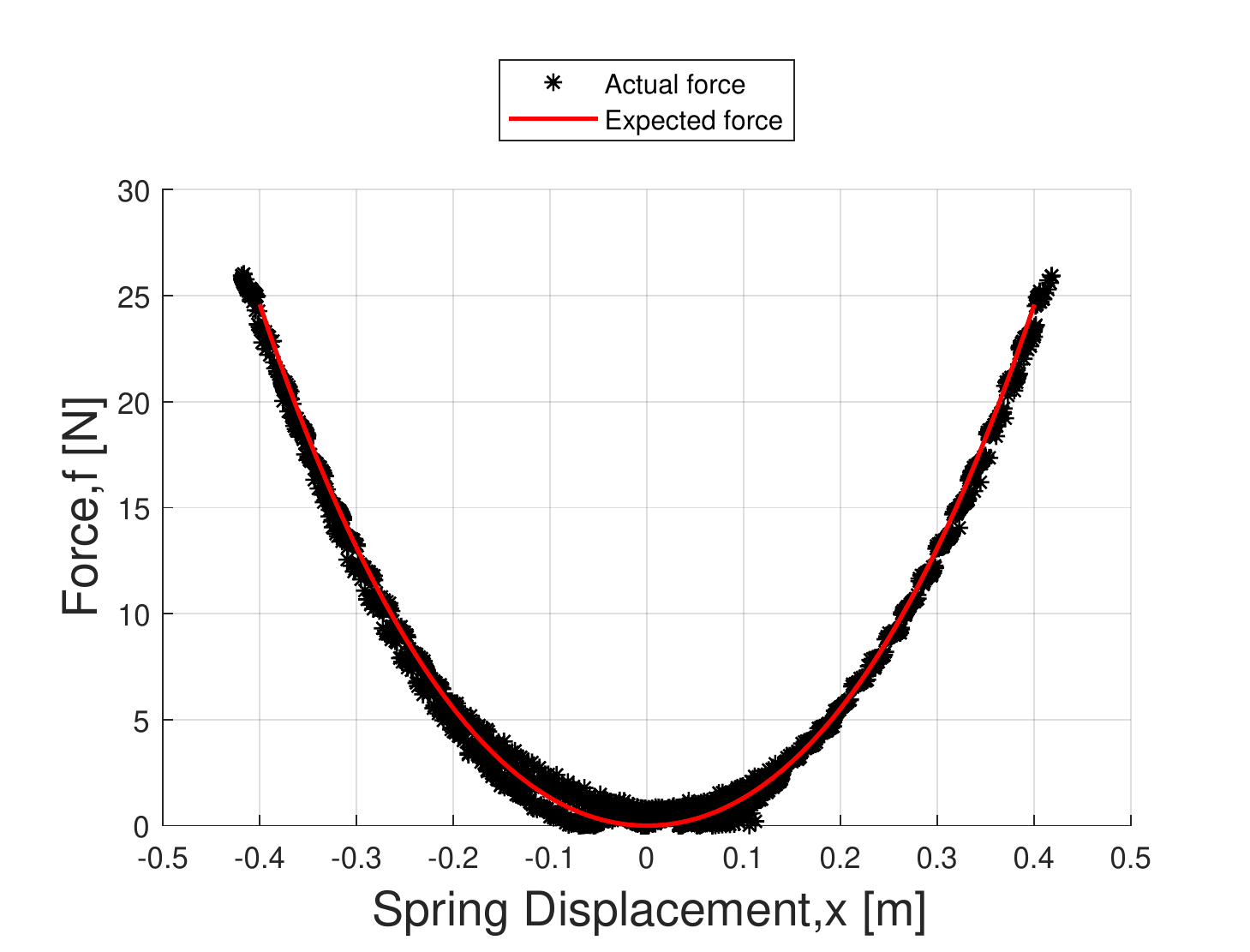}
    \end{minipage}
    }
    \caption{Virtual spring characteristics. Given the virtual linear spring $K_{vs}=75N/m$ and the nonlinear spring $K_{vs}=125(e^{|x_{des}^r-x^r|}-1) N/m$. ($x^r$ means the radial distance of the position)}
    \label{spring}
\end{figure}

After compensating, the desired spring characteristics are obtained. Figure 6(a) and Figure 6(b) show the characteristics of a linear spring and an exponential nonlinear spring, respectively. The abscissa coordinates are radial position deviation and the ordinate coordinates are output force. The set spring elastic coefficient in the experiment is relatively small because smaller spring characteristics can amplify and prominently show the tracking error. The virtual springs show ideal coefficient tracking characteristics.

For linear and nonlinear virtual spring characteristics, impact mitigation performance and position tracking performance should be realized simultaneously. The leg locomotion is divided into two phases: the stance phase and the swing phase. The impact happens in the stance phase and should be effectively mitigated, and the foot of the robot should have ideal position tracking performance in the swing phase. Although linear spring compliance also has impact mitigation capability \cite{Dong2014High} \cite{Semini2015Towards}, nonlinear virtual spring compliance has better performance, based on which we presented nonlinear compliance control on quadruped robots. Considering the leg as a spring-damper mechanism, the dynamic process after the impact is similar to a second-order system impact response. The knee load (similar to the force of spring) reaches a peak firstly, then falls back and fluctuates, and eventually stabilizes at the gravity load. Impact mitigation characteristic requires that the impact peak of the load should be small, also the change of the robot's COM caused by the impact should be small.

Reducing the stiffness of the virtual spring $K_{vs}$ means a smaller impact peak and fewer load fluctuations. However, low stiffness increases the position tracking error and causes the COM position to change more. Inversely, high stiffness reduces the position tracking error, and the COM position change with a greater impact peak and more load fluctuations. After the impact occurs, $K_{vs}$ should be actively reduced to alleviate the impact peak, and after the impact peak, $K_{vs}$ should be actively increased to maintain the stability of COM. High $K_{vs}$ should be kept in the swing phase to guarantee the position tracking performance. A feasible scheme can be expressed as

\begin{small}
\begin{equation}
K_{vs}=K_{vs1}+\frac{2K_{vs2}}{exp(K_{cv}(x_{des}^r-x^r))+exp(-K_{cv}(x_{des}^r-x^r))}
\end{equation}
\end{small}
where $K_{cv}$ is a constant. $K_{vs1}$ and $K_{vs2}$ are constants that describe the range of $K_{vs}$, $K_{vs}$ varies between $K_{vs1}$ and $(K_{vs1}+K_{vs2})$. After the impact occurs, the radial position deviation $(x_{des}^r-x^r)$ increases significantly, $K_{vs}$ is actively reduced. After the impact peak, the COM position begins to recover and  $K_{vs}$ is actively increased. In the swing phase $(x_{des}^r-x^r)$ is small, and high $K_{vs}$ is maintained automatically.

Leg locomotion state is divided into compression state and extension state after the impact. The leg is compressed to alleviate the impact peak and then extended to restore the COM position. The radial velocity error $(\dot{x}_{des}^r-\dot{x^r})$ increases greatly in the compression state, after which the controller will produce a large reverse torque $\tau_{ref}$ to compensate for this deviation. Therefore, $K_d$ should be actively reduced in the compression state, which means a smaller impact peak. $K_d$ should be actively increased in the extension state to restore the COM position. The adjustment of $K_d$ is related to leg locomotion state, a feasible scheme is based on the error function

\begin{equation}
K_d=K_{d1}+K_{d2}\frac{1+{\rm erf}(-K_{cd}(\dot{x}_{des}^r-\dot{x^r}))}{2}
\end{equation}
where ${\rm erf}(x)=\frac{2}{\sqrt{\pi}}\int_{0}^{x}e^{-{\eta}^2}{\rm d}{\eta}$ and $K_{cd}$ is a constant. $K_{d1}$ and $K_{d2}$ are constants that describe the range of $K_d$, $K_d$ varies between $K_{d1}$ and $(K_{d1}+K_{d2})$. In the swing phase $K_d\approx K_{d1}+K_{d2}/2$.

Although the load distribution of the robot leg joint is related to the joint configuration, the knee joint bears the main impact during the robot locomotion. The active adjustment of parameter $K_d$ is based on velocity error $(\dot{x}_{des}^r-\dot{x^r})$, which is effective in one direction. Therefore, the control scheme described in this subsection can also be applied to the swing hip joint, but it can not be directly applied to the span hip joint. For the span hip joint, the impact direction may come from left or right directions, not from one direction, so the parameter $K_d$ should be a constant. The active adjustment of the parameter $K_{vs}$ is independent of the impact direction. The scheme can be directly applied to the span hip joint.

\subsection{Stability analysis}

The stability of the closed-loop control system incorporating the active compliance/impedance controller is  investigated. Although the motor-leg model is a third-order linear system, the active compliance/impedance controller is nonlinear and time-varying. $K_{vs}$ varies between $K_{vs1}$ and $(K_{vs1}+K_{vs2})$, while $K_d$ varies between $K_{d1}$ and $(K_{d1}+K_{d2})$. The stability analysis plays a guiding role for parameters $K_{vs1},K_{vs2},K_{d1},K_{d2}$. For the combination of nonlinear controller and linear system, the Popov stability criterion is a reasonable method. Assuming that $K_d$ is a constant and $K_{vs}$ is a nonlinear gain, the stability limit of $K_{vs}$ with respect to $K_d$ can be found. Firstly, the closed-loop control system incorporating the active compliance/impedance controller is restructured. By changing the output, the constant $K_d$ and the derivative element are added to the motor-leg model (eq. (2)) and become a new linear transfer function $P(s)$, which is described by

\begin{eqnarray}
P(s)&=&\frac{\frac{K_w}{(K_w+J_ls^2+B_ls)(J_ms+B_m)+K_w(J_ls+B_l)}}{s\left[1+K_d\frac{K_w}{(K_w+J_ls^2+B_ls)(J_ms+B_m)+K_w(J_ls+B_l)}\right]} \notag \\
    &=&\frac{K_w}{as^4+bs^3+cs^2+ds}
\end{eqnarray}
where

\begin{equation}
\left\{
             \begin{array}{lr}
             a=J_lJ_m   \\
             b=J_mB_l+J_lB_m   \\
             c=J_mK_w+J_lK_w+B_lB_m   \\
             d=K_wB_m+K_wB_l+K_dK_w
             \end{array}
\right.
\end{equation}

The Popov plot $Re\left[P(j\omega)\right]$ related to $\omega$$Im\left[P(j\omega)\right]$ of $P(j\omega)$ is used to analyse the stability, where $\omega$ is the frequency, $Re\left[P(j\omega)\right]$ is the real part of $P(j\omega)$ and $Im\left[P(j\omega)\right]$ is the imaginary part of $P(j\omega)$. The Popov stability criterion is a  sufficient condition: if the Popov plot of $P(j\omega)$ lies to the right of the Popov line completely, the closed-loop system is global asymptotic stability. The Popov line passes through $(-K_{vs,max}^{-1},j0)$, $K_{vs,max}$ is the maximum $K_{vs}$ that guarantees the stability of the closed-loop system ($K_{vs,max}>K_{vs}>0$). $Re\left[P(j\omega)\right]$ and $\omega$$Im\left[P(j\omega)\right]$ are calculated by

\begin{equation}
\left\{
             \begin{array}{lr}
             Re\left[P(j\omega)\right]=\frac{K_w(a\omega^4-c\omega^2)}{(a\omega^4-c\omega^2)^2+\omega^2(d-b\omega^2)^2}\\
             {\omega}Im\left[P(j\omega)\right]=\frac{-K_w{\omega^2}(d-b\omega^2)}{(a\omega^4-c\omega^2)^2+\omega^2(d-b\omega^2)^2}
             \end{array}
\right.
\end{equation}

Let ${\omega}Im\left[P(j\omega_0)\right]$=0, the Popov stability criterion is described by

\begin{equation}
\left\{
             \begin{array}{lr}
              \omega_0=\sqrt{\frac{d}{b}}\\
             Re\left[P(j\omega_0)\right]=\frac{K_w}{\frac{ad^2}{b^2}-\frac{cd}{b}}\\
             K_{vs,max}=-\frac{1}{ReP(j\omega_0)}
             \end{array}
\right.
\end{equation}

From eq. (16), $K_{vs,max}$ is calculated by

\begin{eqnarray}
K_{vs,max}&=&\frac{(J_mK_w+J_lK_w+B_lB_m)(B_m+B_l+K_d)}{J_mB_l+J_lB_m}\notag\\
&-&\frac{J_mJ_lK_w(B_m+B_l+K_d)^2}{(J_mB_l+J_lB_m)^2}
\end{eqnarray}
where $K_d>0,K_{vs,max}>K_{vs}>0$.

In fact, the same method can be used to calculate the relationship between the limitation of $K_d$ and $K_{vs}$ (assuming that $K_{vs}$ is a constant and $K_d$ is a nonlinear gain), but its form is much more complex. Equation (17) shows the functional relationship between $K_{vs,max}$ and $K_d$, and the parameters of the controller ($K_{vs1},K_{vs2},K_{d1},K_{d2}$) can be selected reasonably based on this relation. In practical application, due to the inner loop PI controller and various unmodeled characteristics, the parameters selection of active compliance/impedance controller should be more conservative than the theoretical value. The relation between leg stiffness and animal running performance is studied by biologists \cite{Lee2014Scaling}.

\section{Experiments}

A series of experiments are performed with the quadruped robot SCIT Dog that used only active compliance and impedance. Three experimental results will be shown to illustrate the performance of the proposed controller after the friction coefficient is determined experimentally. The experiment shows that SCIT Dog has a good ability to interact with the ground. Nvidia Xavier onboard processor is the main control unit. The data is transmitted through the UDP/IP network and CAN bus to implement the motor control. Supported by hardware, a sufficient torque control frequency is realized. The parameters and the experiment hardware of the robot are shown in Table 1.

\begin{tiny}
\begin{table}[htbp]
	\centering
	\caption{Robot experiment parameters.}
	\begin{tabular}{cccc}
		\toprule  
		Parameters&Definition&Value \\
		\midrule  
        $L\times W\times H$&The robot size&$76cm\times 26cm\times 35cm$\\
        $l_1$&Thigh equivalent length&$22.5cm$\\
        $l_2$&Lower leg equivalent length&$12.5cm$\\
        $M$&Total mass of robot&$15kg$\\
        &Perception sensors&$IMU$\\
        &Onboard computer&$Nvidia Xavier$\\
        &maximum joint torque&$33Nm$\\
		\bottomrule  
	\end{tabular}
\end{table}
\end{tiny}

\subsection{PI torque controller performance}

The actual control frequency of the PI controller in the inner loop of the system reaches 5 $kHz$, which is 5 times the control frequency of the outer loop controller (1 $kHz$). Each motor is equipped with a micro-controller as the processing core of the PI controller and the control core of the motor encoder. The structure of the PI controller is shown in Figure 4. The cutoff frequency of the first-order low-pass filter is 20Hz. To demonstrate the performance of the PI controller, the robot joint is fixed in this experiment. Since the torque sensor measures the external torque of the joint, the relationship between the expected torque and the actual torque cannot be compared during joint rotation. Given an expected sinusoidal torque signal of a specific frequency, the experiment result is shown in Figure 7.

\begin{figure}[!ht]
\centering
\includegraphics[width=3.0in]{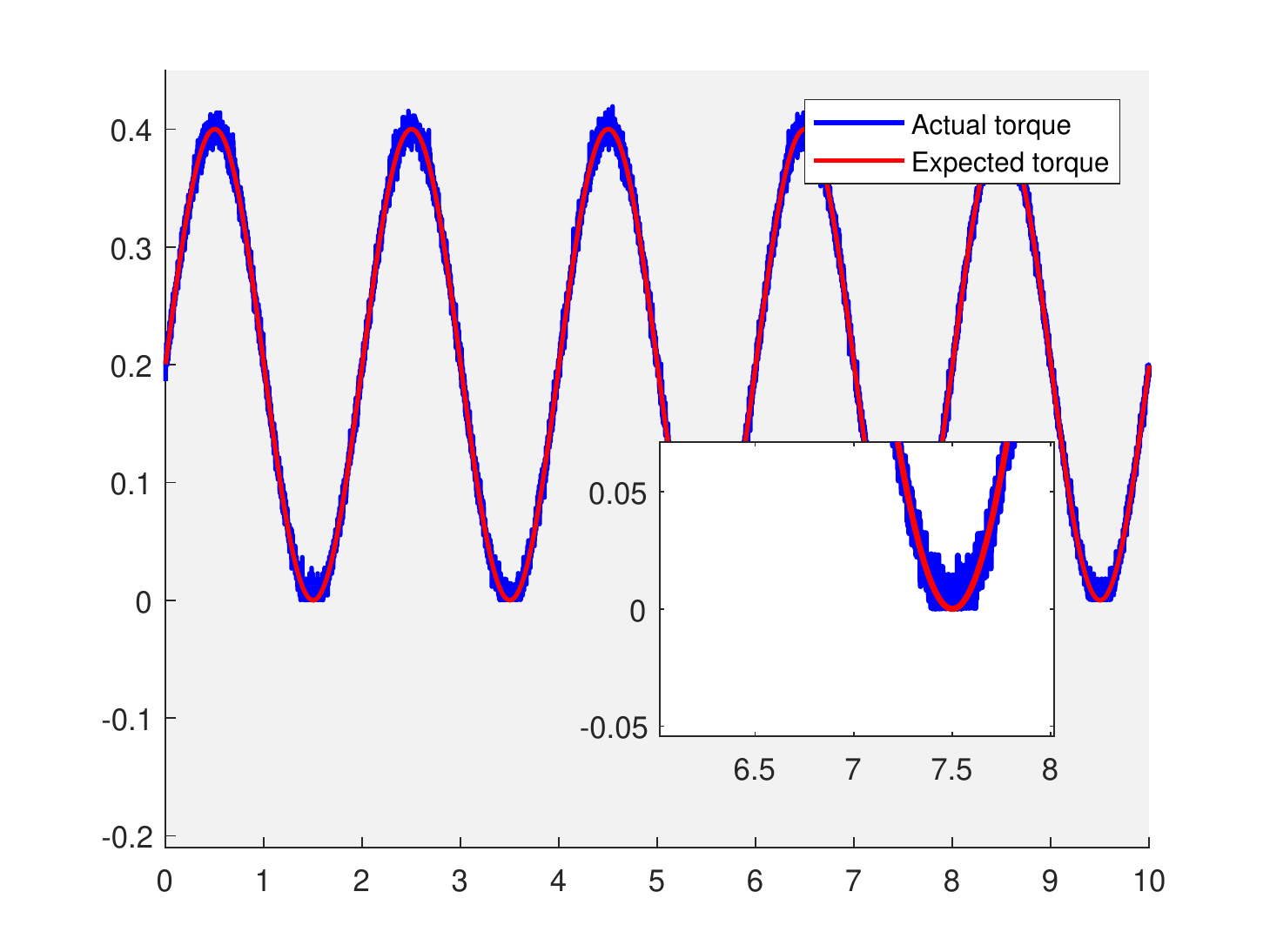}%
\caption{PI torque controller performance.}
\label{fig_sim}
\end{figure}

Although the output torque has oscillation, the PI torque controller still shows good tracking performance, and the oscillation of the PI controller in the inner loop hardly affects the outer loop controller.

\subsection{Position tracking experiment}

According to the analysis in Section 3, Subsection C, the performance of position tracking needs to be concerned in the swing phase. In this experiment, the robot leg is suspended, meaning that there are only gravity and no ground reaction force. (The control indicators to be concerned in the stance phase will be discussed in Section 4, Subsection C.) A desired radial sinusoidal position signal is applied to the foot and the actual position of the foot is measured. The position tracking performances are shown in Figure 8. Blue curves are the expected position and the red curves are the actual position.

\begin{figure}[!ht]
    \centering
    \subfigure[Linear spring position tracking.]{
    \begin{minipage}[b]{0.45\textwidth}
    \centering
    \includegraphics[width=2.5in]{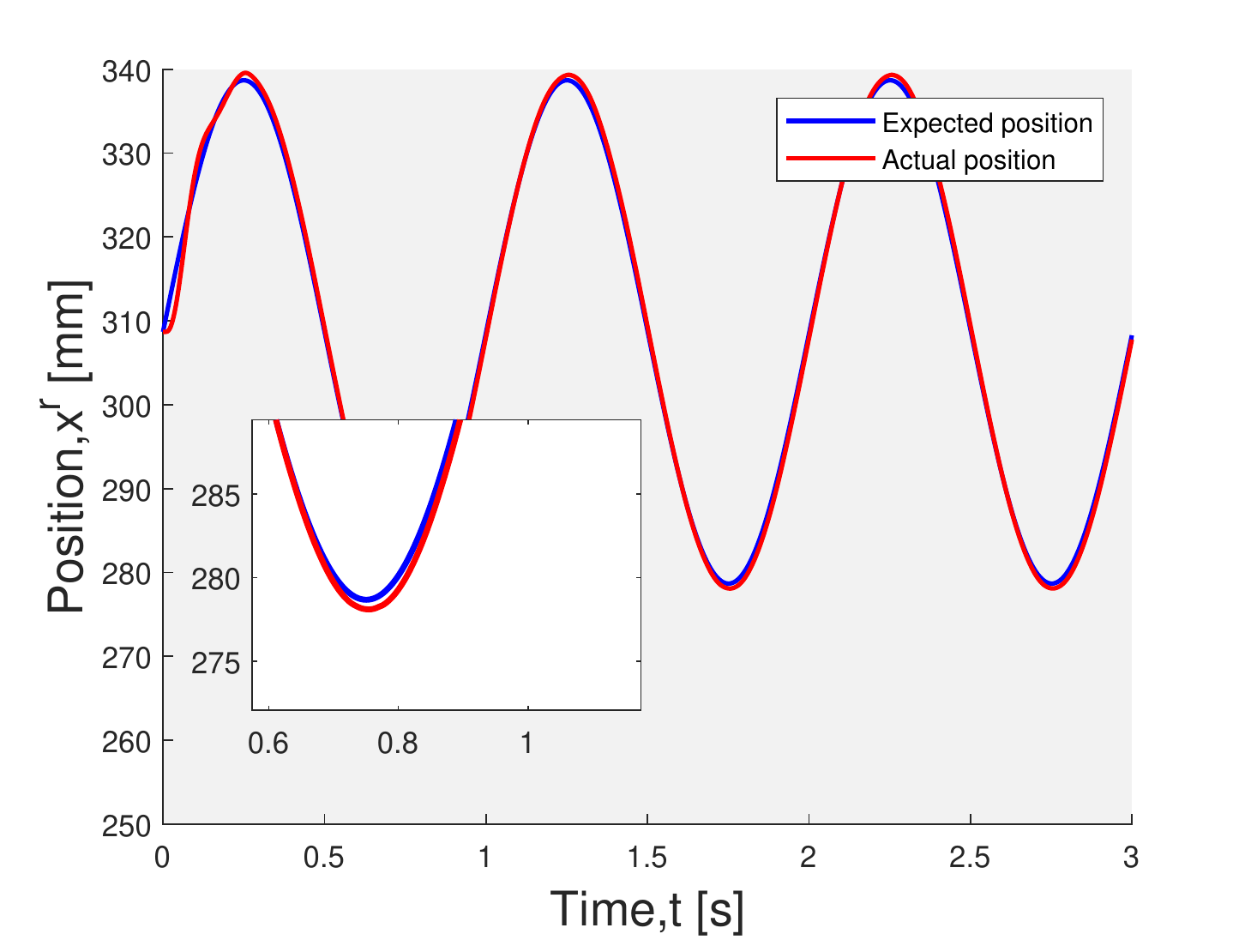}
    \end{minipage}
    }
    \subfigure[Nonlinear spring position tracking.]{
    \begin{minipage}[b]{0.45\textwidth}
    \centering
    \includegraphics[width=2.5in]{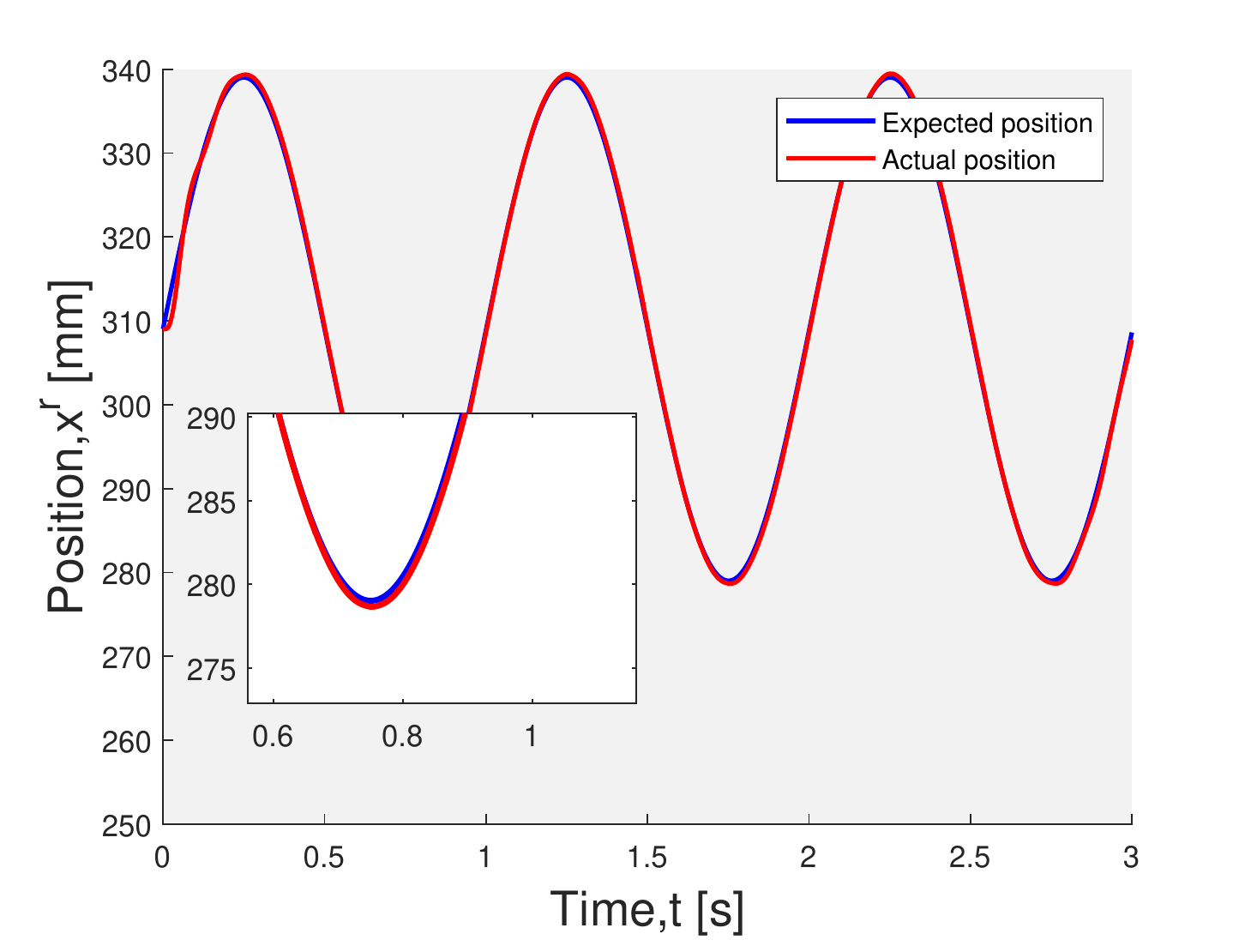}
    \end{minipage}
    }
     \subfigure[Position tracking deviation.]{
    \begin{minipage}[b]{0.45\textwidth}
    \centering
    \includegraphics[width=2.5in]{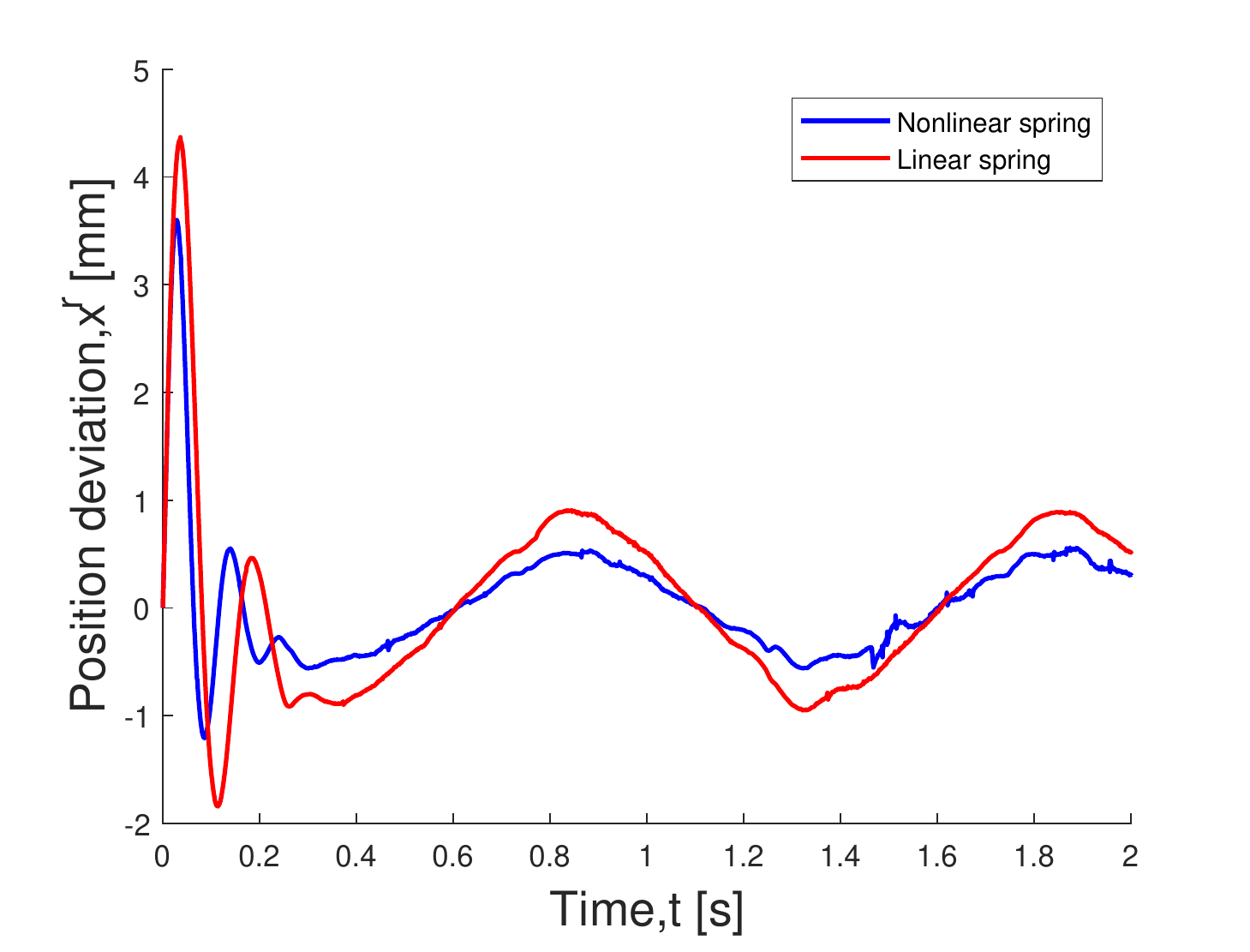}
    \end{minipage}
    }
    \caption{Position tracking experiment.}
    \label{spring}
    \end{figure}

The experimental results show that the nonlinear spring ($K_{vs1}=500N/m,K_{vs2}=500N/m$) has better position tracking performance than the linear spring ($K_{vs}=750N/m$). The position tracking error of the linear spring is $\pm 4.4mm$ and the position tracking error of the nonlinear spring is $\pm 3.6mm$ (Figure 8 (c)). In the swing phase, high $K_{vs}$ of nonlinear spring is maintained automatically.

\subsection{Impact mitigation experiments}

In the experiments, the robot stands with all four legs on the ground. A vertical impulse (200 $N$ peak, 500 $N$ peak, 800 $N$ peak and 200 ms duration) is then applied to the robot through the feed-forward controller. The robot jumps up and then impacts on the ground. When all four feet touch the ground, it is regarded as the beginning of impact. The COM position is estimated through the leg kinematics. The experiment results are shown in Figure 9-11.

The performances of linear spring (four kinds) and nonlinear spring are compared. Linear springs are: 1) $K_{vs}=1000N/m$, $K_d=50Ns/m$ (high stiffness and low damping, red curves) 2) $K_{vs}=1000N/m$, $K_d=150Ns/m$ (high stiffness and high damping, green curves) 3) $K_{vs}=500N/m$, $K_d=50Ns/m$ (low stiffness and low damping, blue curves) 4) $K_{vs}=500N/m$, $K_d=150Ns/m$ (low stiffness and high damping, yellow curves). Nonlinear spring is : $K_{vs}=variable (500\sim1000N/m)$, $K_d=variable (50\sim150Ns/m)$ (the black curves). From Figure 9(b),10(b),11(b),  it can be seen that a higher joint stiffness $K_{vs}$ reduces the steady-state error of the COM vertical position (created by gravity forces). But a higher joint stiffness $K_{vs}$ reduces a larger knee load (Figure 9(a),10(a),11(a)). The experiment results show that the nonlinear spring has a better impact mitigation performance (smaller knee load and smaller COM change). From Figure 9(c),10(c),11(c) and Figure 9(d),10(d),11(d), after the impact occurs, the active compliance/impedance controller reduces $K_{vs}$ and $K_d$ to alleviate the impact peak. After the impact peak, $K_{vs}$ and $K_d$ is actively increased to maintain the stability of COM. Various linear springs show disadvantages, such as larger knee loads or COM deviations. High linear $K_{vs}$ and $K_d$ cause larger average knee load, while low linear $K_{vs}$ and $K_d$ cause more COM change in vertical direction. The experimental results confirm the analysis in Section 3 and demonstrate that the active compliance/impedance controller proposed in this paper has a better mitigation capability for impact.

\begin{figure}[t]
    \centering
    \subfigure[Average knee load.]{
    \begin{minipage}[b]{0.5\textwidth}
    \centering
    \includegraphics[width=3.5in]{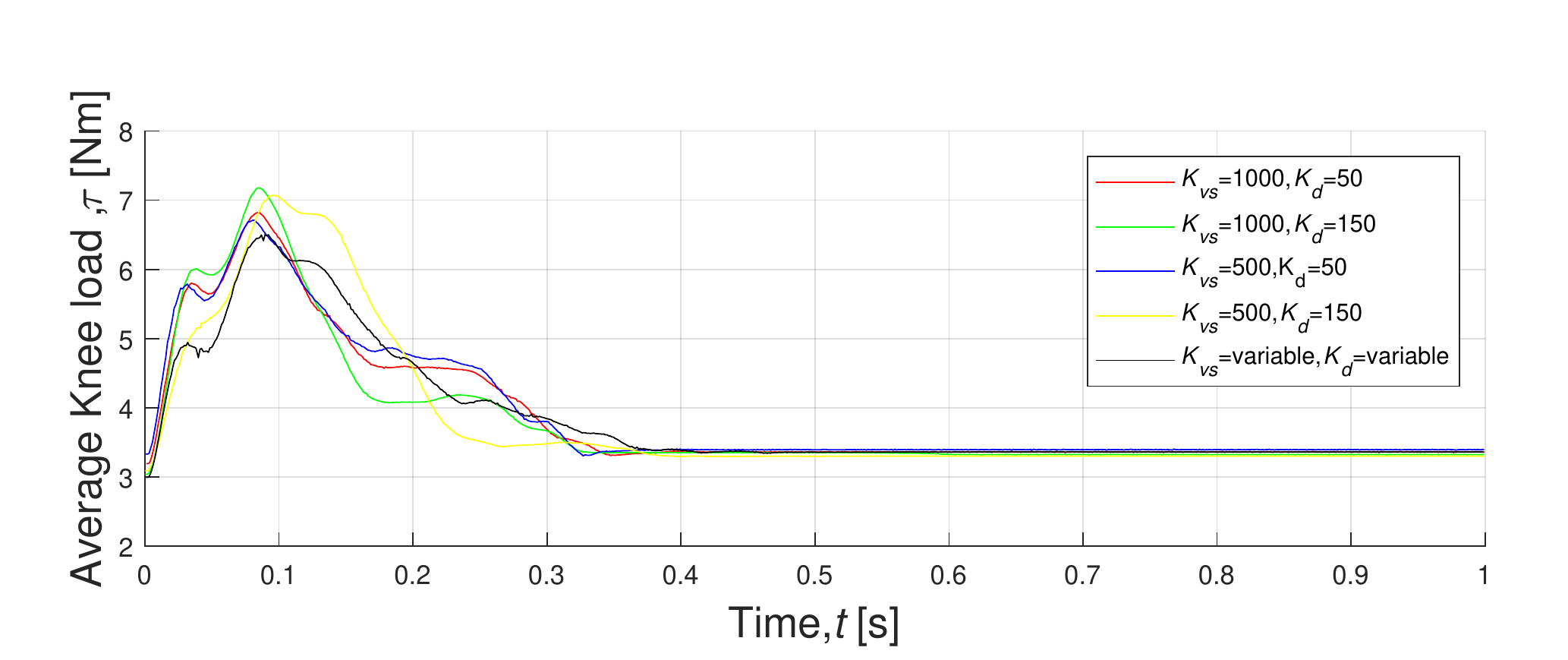}
    \end{minipage}
    }
    \subfigure[Vertical direction COM deviation.]{
    \begin{minipage}[b]{0.5\textwidth}
    \centering
    \includegraphics[width=3.5in]{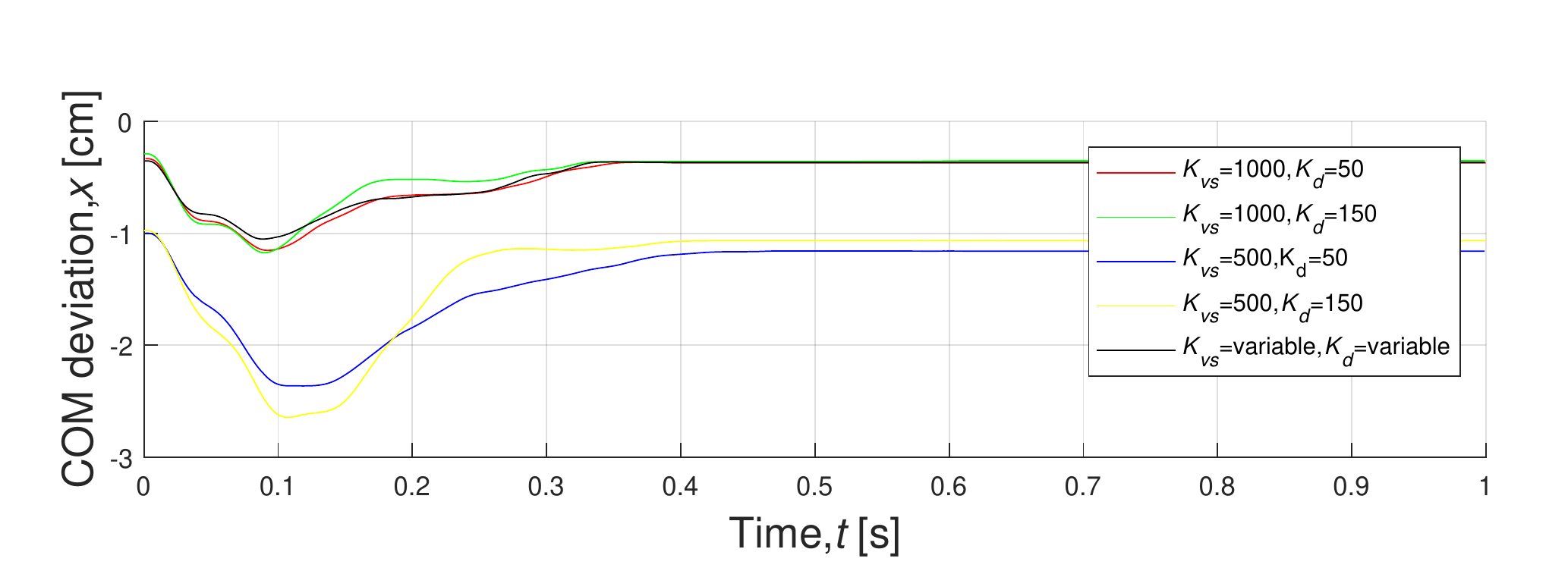}
    \end{minipage}
    }
    \subfigure[Time series of $K_{vs}$.]{
    \begin{minipage}[b]{0.5\textwidth}
    \centering
    \includegraphics[width=3.5in]{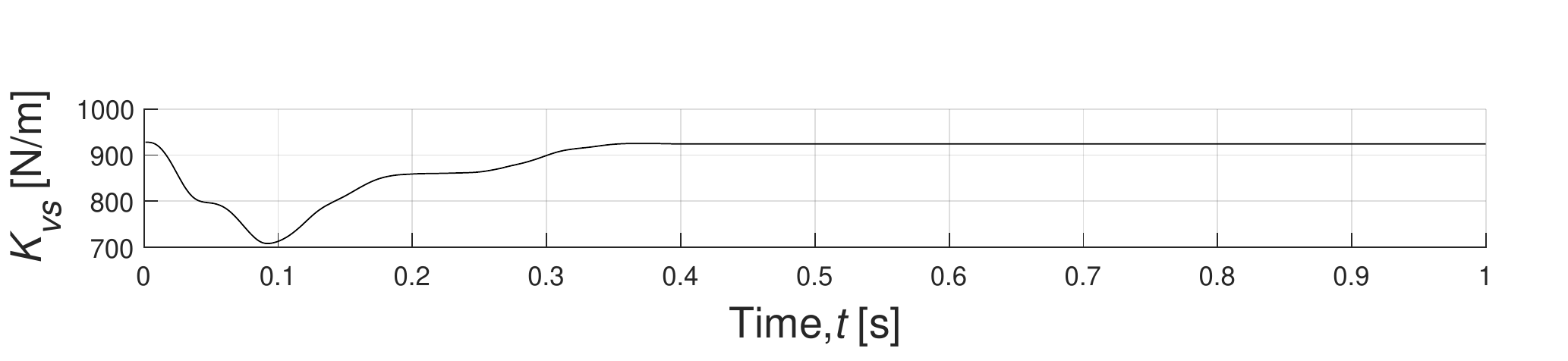}
    \end{minipage}
    }
    \subfigure[Time series of $K_d$.]{
    \begin{minipage}[b]{0.5\textwidth}
    \centering
    \includegraphics[width=3.5in]{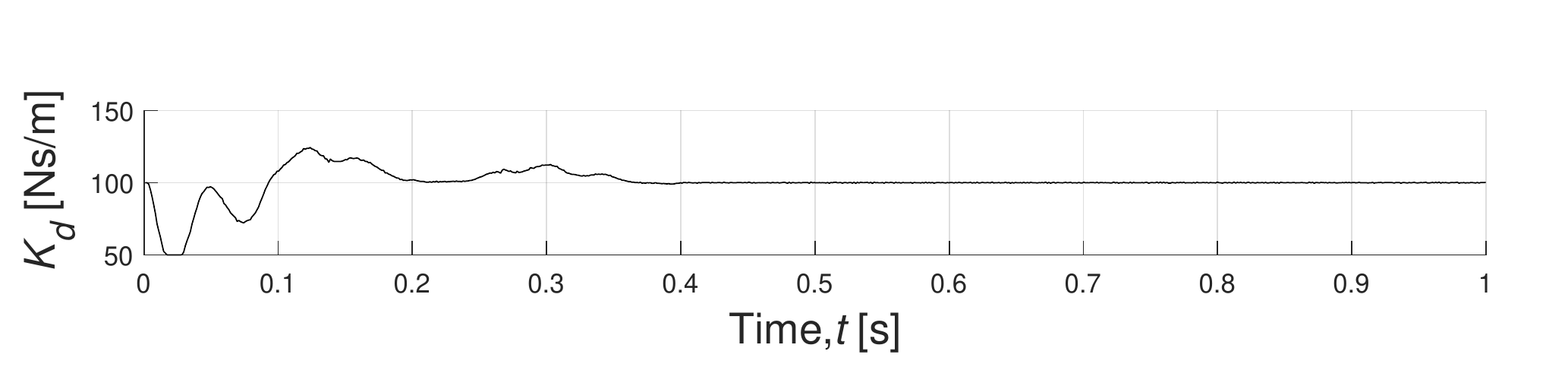}
    \end{minipage}
    }
    \caption{Impact mitigation experiments (Feed-forward vertical impulse 200 $N$ peak and 200 ms duration).}
    \label{spring}
    \end{figure}

\begin{figure}[t]
    \centering
    \subfigure[Average knee load.]{
    \begin{minipage}[b]{0.5\textwidth}
    \centering
    \includegraphics[width=3.5in]{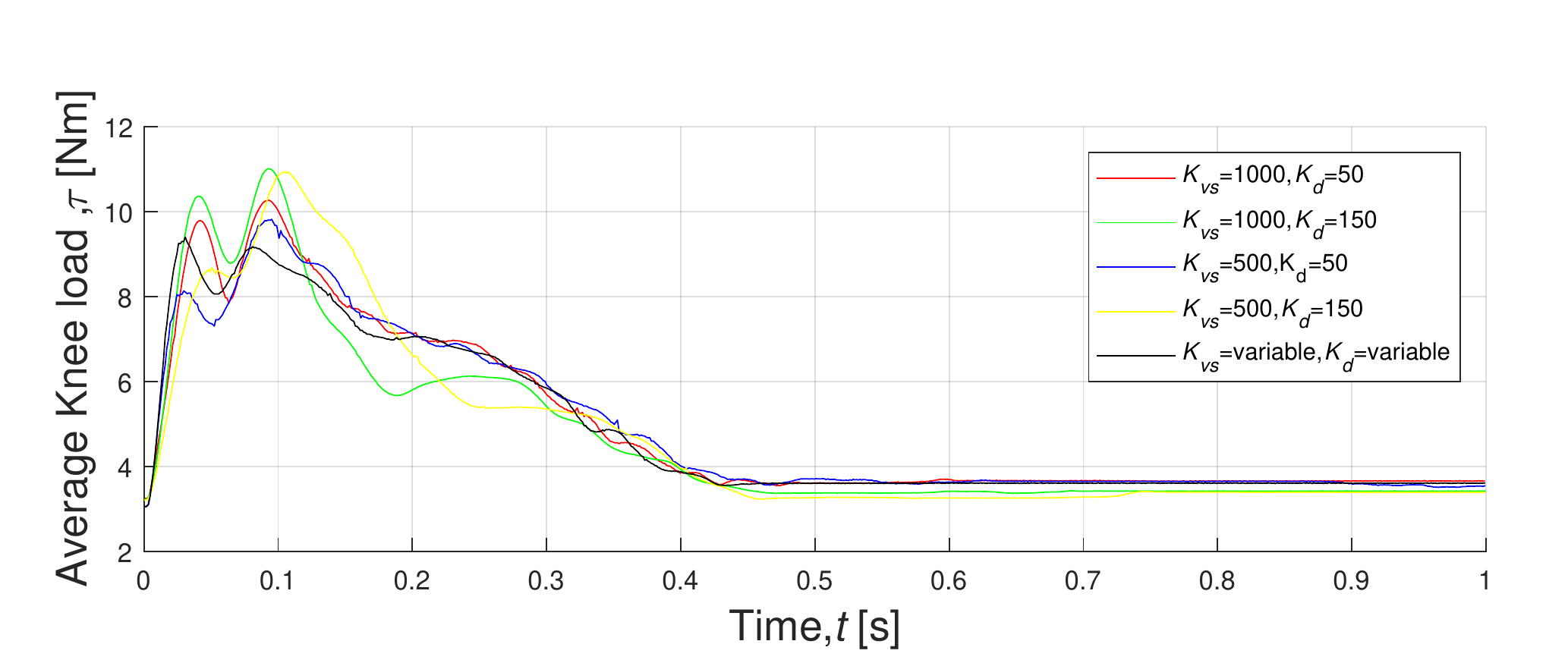}
    \end{minipage}
    }
    \subfigure[Vertical direction COM deviation.]{
    \begin{minipage}[b]{0.5\textwidth}
    \centering
    \includegraphics[width=3.5in]{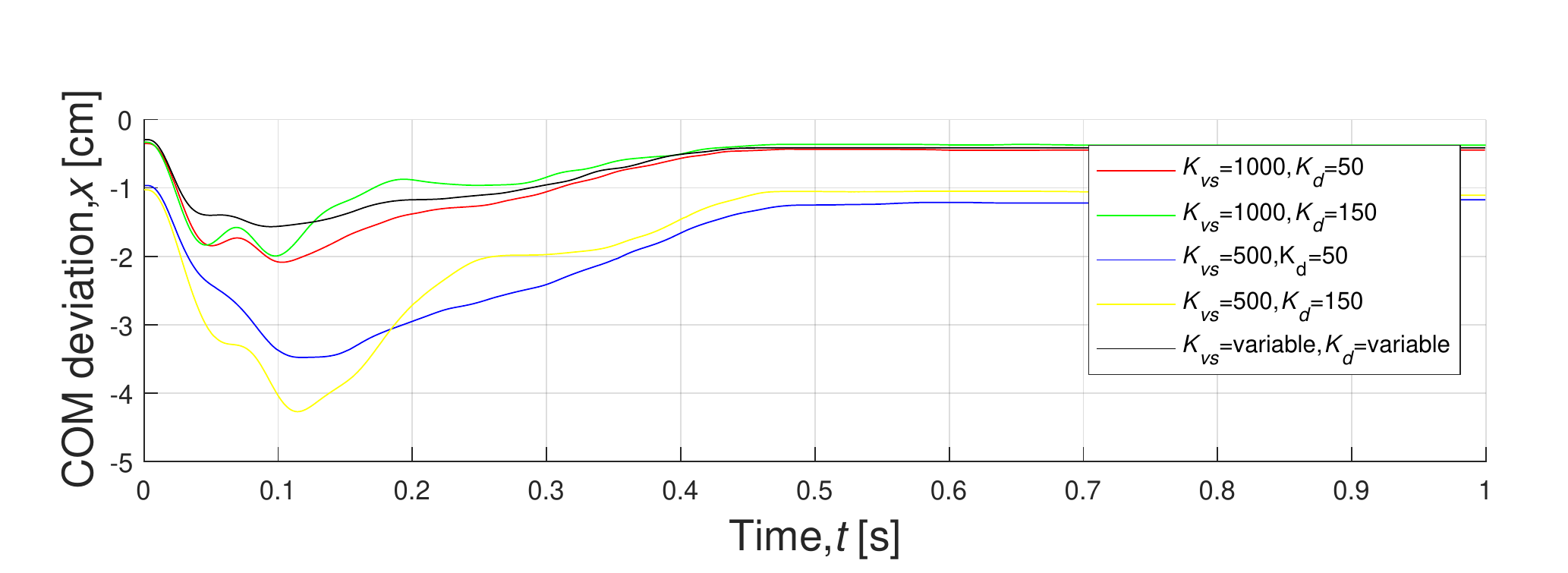}
    \end{minipage}
    }
    \subfigure[Time series of $K_{vs}$.]{
    \begin{minipage}[b]{0.5\textwidth}
    \centering
    \includegraphics[width=3.5in]{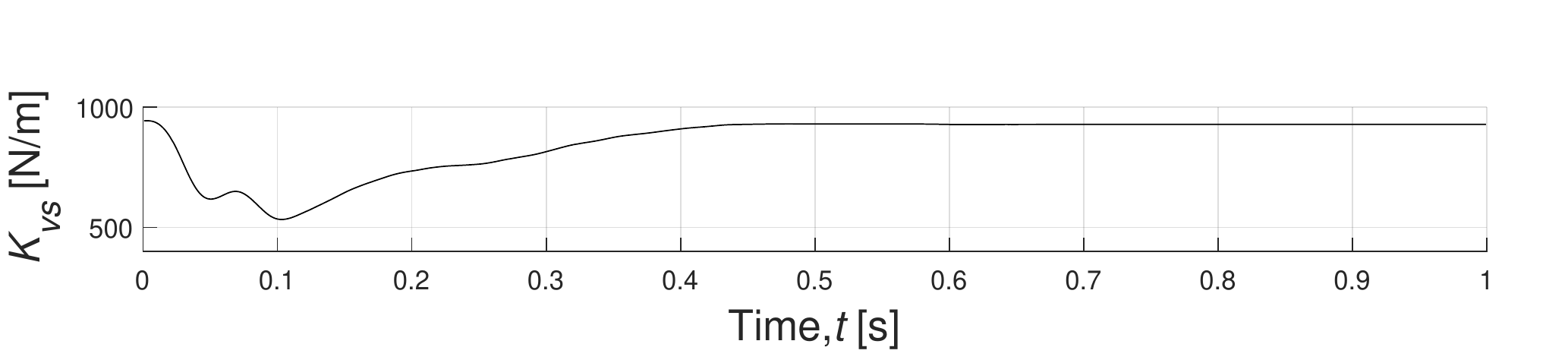}
    \end{minipage}
    }
    \subfigure[Time series of $K_d$.]{
    \begin{minipage}[b]{0.5\textwidth}
    \centering
    \includegraphics[width=3.5in]{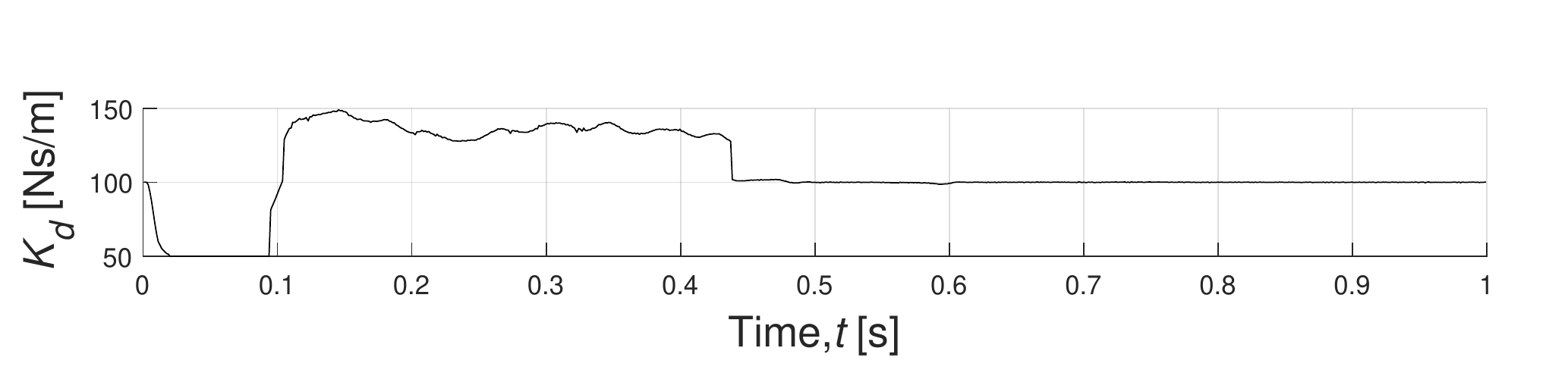}
    \end{minipage}
    }
    \caption{Impact mitigation experiments (Feed-forward vertical impulse 500 $N$ peak and 200 ms duration).}
    \label{spring}
    \end{figure}

\begin{figure}[t]
    \centering
    \subfigure[Average knee load.]{
    \begin{minipage}[b]{0.5\textwidth}
    \centering
    \includegraphics[width=3.5in]{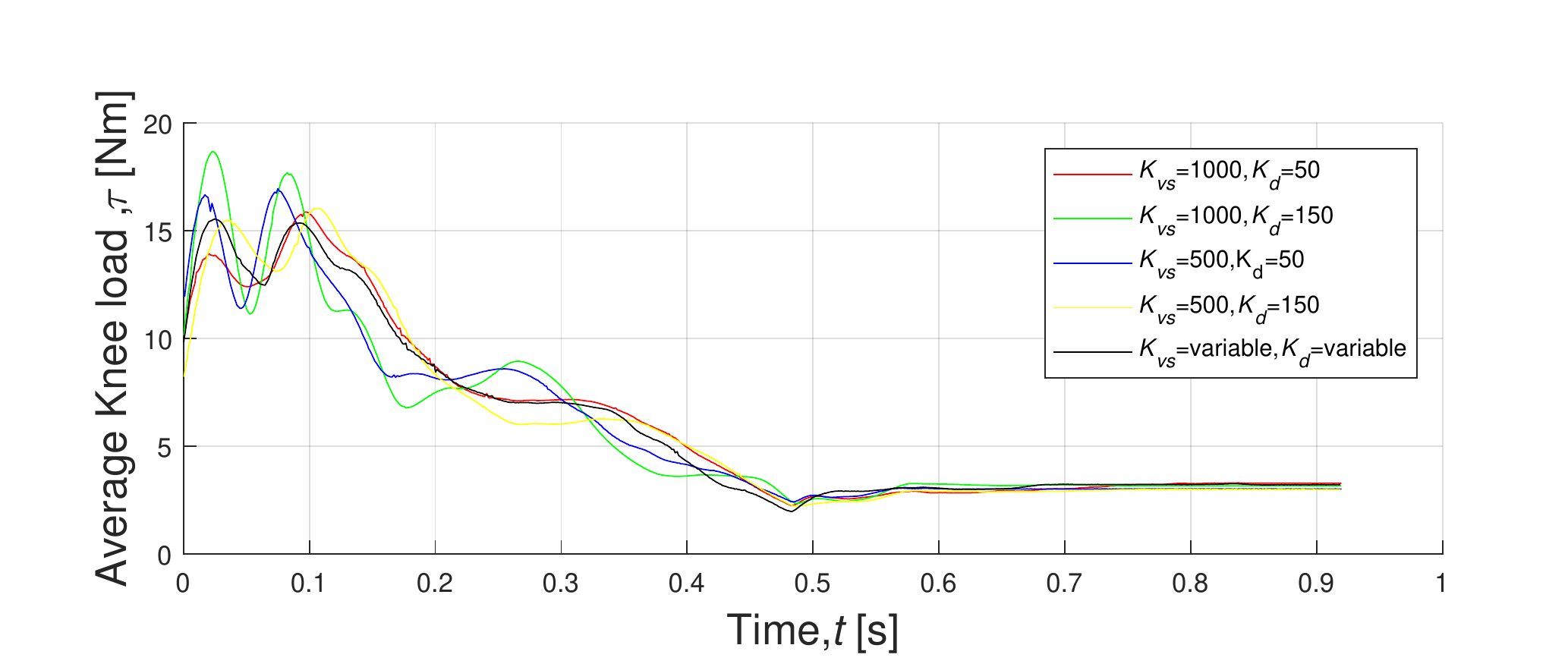}
    \end{minipage}
    }
    \subfigure[Vertical direction COM deviation.]{
    \begin{minipage}[b]{0.5\textwidth}
    \centering
    \includegraphics[width=3.5in]{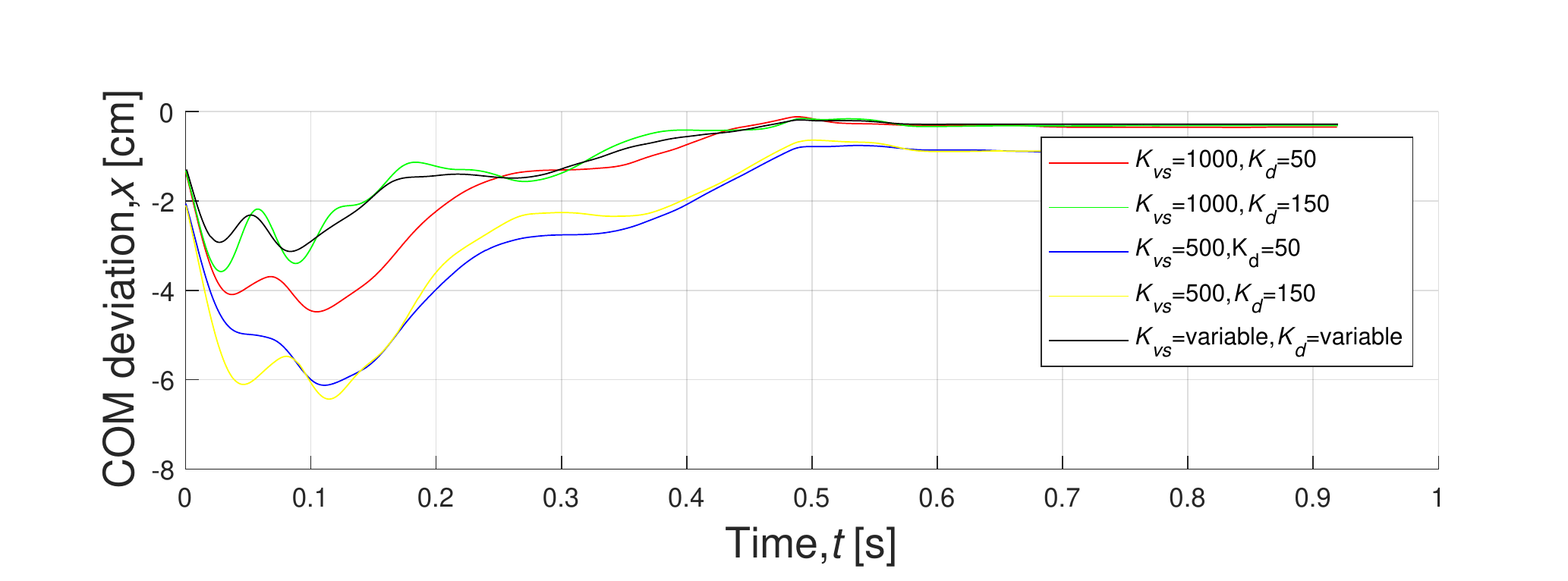}
    \end{minipage}
    }
    \subfigure[Time series of $K_{vs}$.]{
    \begin{minipage}[b]{0.5\textwidth}
    \centering
    \includegraphics[width=3.5in]{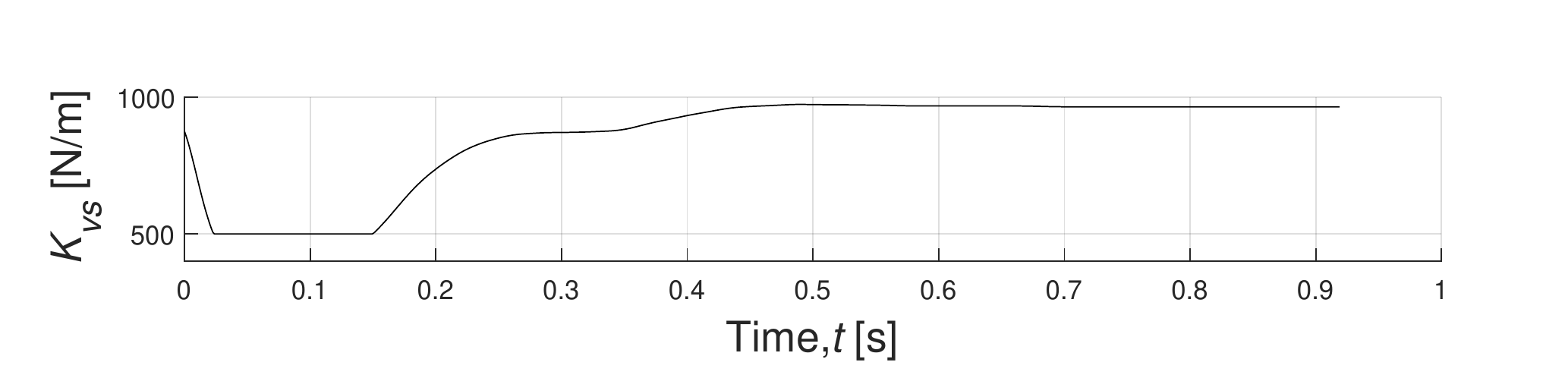}
    \end{minipage}
    }
    \subfigure[Time series of $K_d$.]{
    \begin{minipage}[b]{0.5\textwidth}
    \centering
    \includegraphics[width=3.5in]{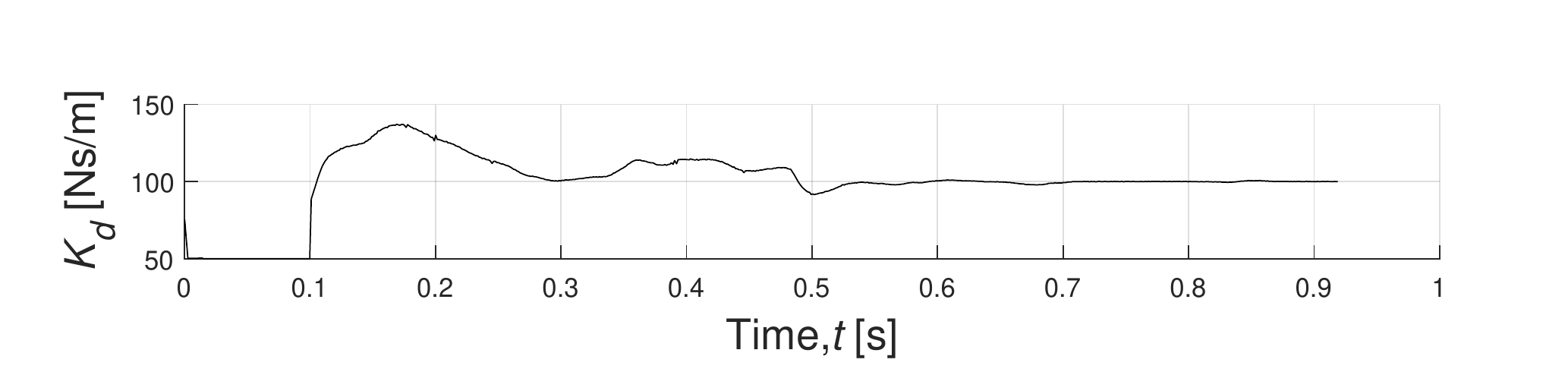}
    \end{minipage}
    }
    \caption{Impact mitigation experiments (Feed-forward vertical impulse 800 $N$ peak and 200 ms duration).}
    \label{spring}
    \end{figure}

\section{Conclusions}

{\label{687807}}

A new nonlinear active compliance control is applied to quadruped robots and achieves better interaction capability than linear compliance control. Experiments show that SCIT Dog's legs have the ability to cope with large and impulsive impact forces in running and jumping without any passive compliance mechanism. This work also shows a reasonable control scheme for leg dynamic locomotion. A PI controller is used to directly control the motor current to solve the nonlinear characteristics of motor voltage input and torque output. The model-based compensation method is successfully implemented on SCIT Dog. Active compliance control and active impedance control have been applied, and experiments show both ideal impact mitigation performance and position tracking performance. The proposed scheme in this paper can be extended to all legged robots.

\ifCLASSOPTIONcaptionsoff
  \newpage
\fi



%
\bibliography{biblio}{}
\bibliographystyle{IEEEtran}

%




\end{document}